\documentclass[letterpaper]{article} 
\usepackage[preprint]{aaai2027}  
\usepackage[hyphens]{url}  
\usepackage{graphicx} 
\usepackage{natbib}  
\usepackage{caption} 
\usepackage{algorithm}
\usepackage{algpseudocode}

\usepackage{newfloat}
\usepackage{listings}
\DeclareCaptionStyle{ruled}{labelfont=normalfont,labelsep=colon,strut=off} 
\floatstyle{ruled}
\newfloat{listing}{tb}{lst}{}
\floatname{listing}{Listing}

\usepackage{booktabs}

\usepackage{multirow}
\usepackage[table,xcdraw,dvipsnames]{xcolor}
\usepackage{amsmath}
\usepackage{amssymb}
\usepackage{cleveref}

\definecolor{clrFree}{RGB}{0,0,0}
\definecolor{clrObst}{RGB}{76,76,255}
\definecolor{clrStart}{RGB}{255,76,76}
\definecolor{clrGoal}{RGB}{76,255,76}
\definecolor{clrSemanticObst}{RGB}{100,100,255}
\definecolor{clrWaypoint}{RGB}{255,255,76}

\title{FlexPath: Adapting Learned Connectivity Guidance to Path Preferences}

\author{
    Taehyoung Kim\textsuperscript{\rm 1}\equalcontrib,
    Tim Sch\"onbrod\textsuperscript{\rm 1,\rm 2}\equalcontrib,
    David Eckel\textsuperscript{\rm 1},
    Henri Mee{\ss}\textsuperscript{\rm 1}
}

\affiliations{
    \textsuperscript{\rm 1}Fraunhofer IVI
    \qquad
    \textsuperscript{\rm 2}Technische Hochschule Ingolstadt
}

\begin{document}

\maketitle

\begin{abstract}
      Recent learning-based path planners use neural networks to process occupancy representations and approximate heuristics for classical search algorithms, yielding near-optimal paths with reduced search effort. However, these methods are tied to a fixed objective, usually the shortest-path objective, implicit in their supervision. This limits their flexibility to accommodate alternative criteria. 
      We introduce $\textbf{FlexPath}$, a two-stage learned search-guidance framework that first learns a recall-oriented connectivity prior initialized from shortest-path planner demonstrations and then refines this prior using differentiable path-shape objectives, thereby separating demonstration-based learning of $\textbf{connectivity-biased guidance}$ from subsequent $\textbf{objective specific refinement}$. Beyond enabling adaptation to new routing preferences, the two-stage procedure improves standard shortest-path planning itself: on TMP, FlexPath improves optimal-path recovery from 75.0\% to 88.6\% over TransPath while reducing search expansions by 13.8\%. Ablations show that neither prior learning nor objective fine-tuning alone matches the full pipeline; their combination yields the strongest path cost and search efficiency. We further demonstrate the preference adaptation by adapting guidance to non-shortest-path objectives such as obstacle clearance, class-conditioned obstacle clearance and waypoint following. For clearance with $d_{\min}=2$, FlexPath achieves 96.2\% full clearance satisfaction on feasible instances while maintaining low search effort, and it reaches 98.4\% waypoint-following success.
\end{abstract}


\begin{figure}[t]
    \centering
    \includegraphics[width=0.98\columnwidth]{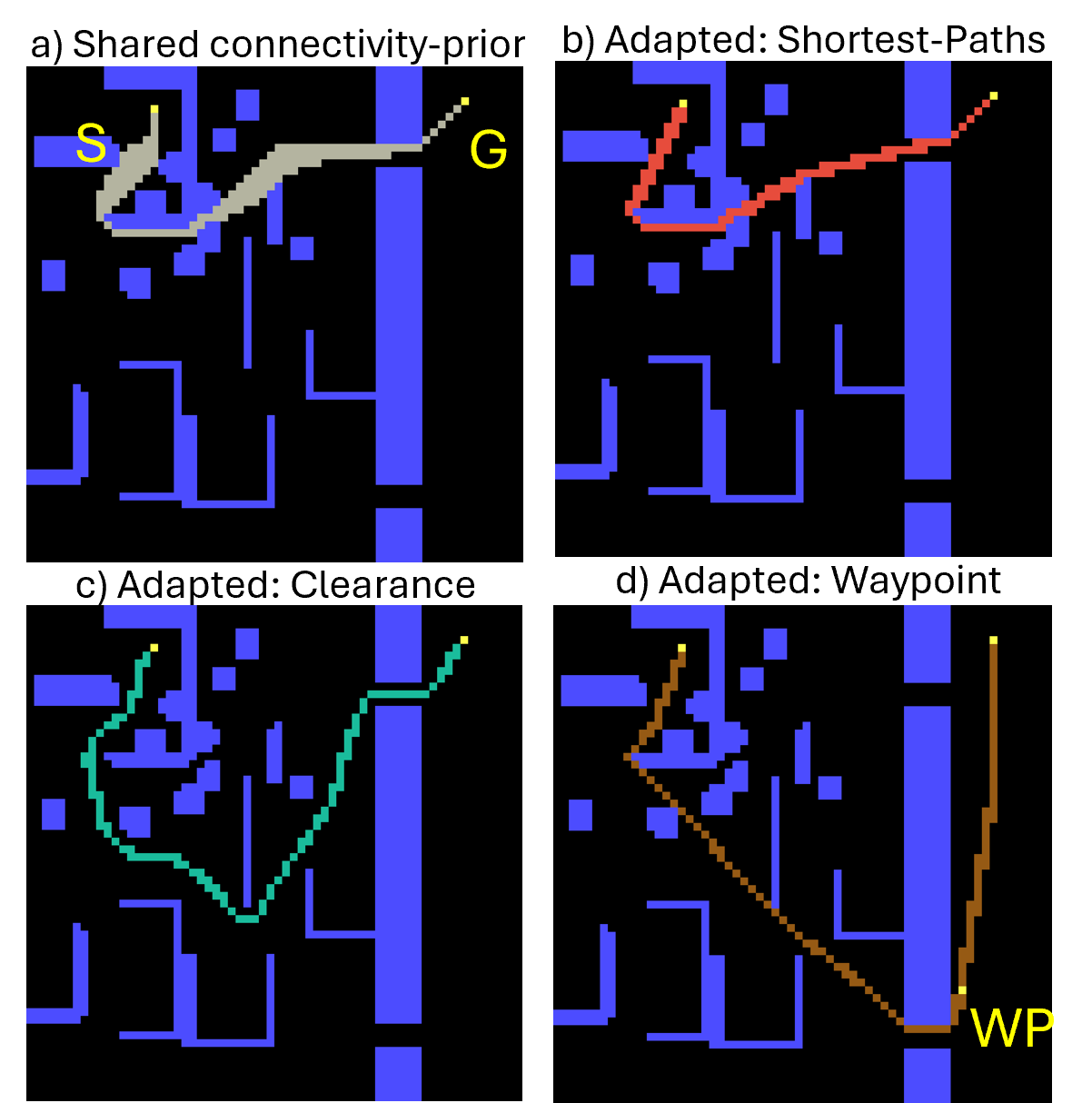}
    \caption{%
        One prior, flexible path adaptation.
        A connectivity soft path mask (a/grey) is fine-tuned via Path Shape Objectives to yield shortest-path (b/red), obstacle-clearance (c/aquamarine), and waypoint-constrained masks (d/brown) (via WP) without retraining from scratch.
    }
    \label{fig:FrontFigure}
\end{figure}

\section{Introduction}
\label{sec:intro}
Path planning is widely used across areas such as mobile robots \cite{MRPP}, autonomous driving \cite{ADPP1,ADPP2}, robot arm manipulation \cite{RAPP1, RAPP2}, and autonomous flight \cite{UAVPP1}. Classical planners like A* \cite{A*} are commonly used as they guarantee valid and optimal solutions under a well-designed cost function. However, as environments get larger and more complex, this approach can lead to exponential growth in complexity, making path planning inefficient or even infeasible. Moreover, real-world navigation requires more than just optimality: systems must account for energy efficiency on resource-constrained platforms, safety via obstacle clearance, and, increasingly, social and semantic constraints as robots operate in human environments \cite{xiao2022motionplanningcontrolmobile}. Existing path-planning frameworks address these problems by designing cost functions or layering cost maps, but this approach is labor-intensive and often struggles to generalize across environments \cite{cost}.

Recent work leverages deep learning to learn heuristics that guide search more efficiently toward near-optimal paths~\cite{bhardwaj2017learningheuristicsearchimitation, yonetani2021pathplanningusingneural, kirilenko2022transpathlearningheuristicsgridbased, iA*, DAA*}. Although these methods made progress to reduce search effort, they rely heavily on expert paths responsible for the planner’s guidance (A* \cite{A*}, Theta* \cite{theta*}, or paths drawn directly by humans). 
This makes it difficult to apply these methods to objectives beyond shortest paths.


Our central hypothesis is that dense learned search guidance should not be trained solely to imitate a final path. Instead, learning a recall-oriented connectivity prior and subsequently optimizing it for the desired path shape can improve the canonical shortest-path task itself. Crucially, the same prior can be refined toward alternative preferences such as clearance, class-conditioned clearance, and waypoint guidance (\Cref{fig:FrontFigure}). On shortest-path planning, this two-stage formulation improves both path quality and search efficiency over prior learned planners, while the ablation shows that neither stage alone matches the full system.

In summary, we introduce \textbf{FlexPath}, a framework that first learns path structure (connectivity) from recall-biased shortest path demonstrations and then refines it to align with task-specific preferences. Our key contributions are:

\begin{itemize}
    \item We introduce an imitation-to-optimization-inspired framework that first learns a reusable, recall-oriented connectivity prior and subsequently refines it through differentiable Path Shape Objectives (PSOs), separating path-structure learning from objective-specific preferences.
    \item We show that prior learning and objective fine-tuning are complementary: their combination improves shortest-path cost, optimal-path recovery and search efficiency over established learned planners.
    \item We demonstrate that the same Stage-1 prior can be adapted to qualitatively different non-shortest-path objectives, including obstacle clearance, class-conditioned avoidance, and waypoint guidance, without additional path labels or retraining the full model from scratch.
\end{itemize}

\section{Related Work}
\label{sec:related}

\noindent{\bfseries Classical Path Planning.} 
Classical path planning is formulated as finding the cost-minimal path between the start and goal states using a discrete search algorithm on a graph or grid representation \cite{RN2020}. Heuristic search algorithms led by A* \cite{A*} provide a strong guarantee of an optimal path under an admissible heuristic, but can become computationally expensive as the search space grows. This motivated the development of suboptimal variants of this algorithm, such as Weighted A* (WA*) \cite{WA*} or Focal Search (FS) \cite{FS}, which trade off path quality and path-finding efficiency. Real-world navigation often requires criteria extended beyond finding such an optimal path, such as maintaining a certain distance from objects \cite{dolgov2008practical} or generating socially acceptable motion \cite{lu2014layered, ColetoTSA}. In the classical framework, implementing such motion requires manually shaping cost functions or layering cost maps, which is labor-intensive and tough to generalize to diverse environments and purposes \cite{choset2005principles}. 

\noindent{\bfseries Learning-Based Path Planning.}
Recent advances in deep learning have led to a wide range of learning-based path planning methods aimed at overcoming the limitations of classical search algorithms \cite{pmlr-v278-xu25a}. Reinforcement learning (RL) approaches formulate path planning as a Markov Decision Process (MDP). Methods such as Q-learning~\cite{ZHOU2024111400, ji2024neuralnetworkdrivenrewardpredictionheuristic, qlearning} and policy-gradient algorithms, such as PPO~\cite{Babu_2023, ppo}, are used to learn a policy via sequential decision-making. Rather than predicting a complete path in a single inference step, these methods produce solutions by iteratively rolling out the learned policy in the environment, collecting trajectories and reward feedback to evaluate performance \cite{ZHANG2024128423}. The sparse and delayed reward signal along sampled trajectories makes single-step supervision of fine-grained geometric details challenging, so their inference often resembles online control rather than direct path prediction~\cite{WU2023115208}. Therefore, such approaches are typically sample-inefficient~\cite{dulacarnold2019challengesrealworldreinforcementlearning, nguyen2025emergencedeepreinforcementlearning}.

A complementary line of work seeks to improve path-planning efficiency through heuristic learning. In this setting, the focus is on instance-independent guidance of classical search: Search as Imitation Learning (SAIL)~\cite{bhardwaj2017learningheuristicsearchimitation} learns a cost-to-go function by imitation of an expert planner, and subsequent work, TransPath \cite{kirilenko2022transpathlearningheuristicsgridbased}, streamlines path planning by predicting Path Probability Maps (PPMs), a spatial guidance representation. Trained with supervised signals from optimal or expert trajectories, these methods bias search towards a high-likelihood region where an expert path is likely to lie. In practice, 2D geometric occupancy or cost maps are used as inputs to predict guidance. Later extensions incorporate additional learned components to further improve efficiency~\cite{THELLIER2026123149}, but continue to rely on supervision derived from optimal trajectories.

Another line of work integrates search directly into neural architectures. Differentiable planners such as Neural A*~\cite{yonetani2021pathplanningusingneural} or iA*~\cite{iA*}, which use differentiable A* formulations, integrate planning dynamics into the training process so gradients can propagate through the search. More recently,  DAA*~\cite{DAA*} further refined this paradigm by incorporating path smoothness objectives. While this tight integration of the search process can improve generalization performance across diverse environments, it requires executing (and often differentiating through) the search during training. Consequently, these methods introduce substantial computational overhead and offer limited flexibility to adjust planning criteria.

\vspace{0.5em}
\noindent{\bfseries Imitation-to-Optimization Paradigm.}
Recent advances in deep learning have demonstrated the effectiveness of this two-stage paradigm. Models first learn a broad structural prior from data through imitation or reconstruction objectives, and are subsequently refined using task- or preference-specific optimization signals. This strategy has been successfully applied in areas such as foundation model alignment~\cite{instructGPT, DPO, GRPO} and policy refinement in robotics~\cite{DAPG, residualpolicylearning}, where an initially learned distribution is steered toward desired behaviors without relearning the underlying structure.

Our method follows an imitation-to-optimization pattern: an initial model is learned from demonstrations, then adapted using objective-specific optimization signals. In FlexPath, imitation provides a connectivity-biased initialization for dense search guidance, while optimization reshapes this guidance according to differentiable path objectives. Unlike preference-learning settings that infer preferences from comparisons or rewards, our second stage uses explicitly specified path-shape functionals.

\section{Methodology}
\label{sec:method}

\begin{figure*}[t]
    \centering
    \includegraphics[width=\textwidth]{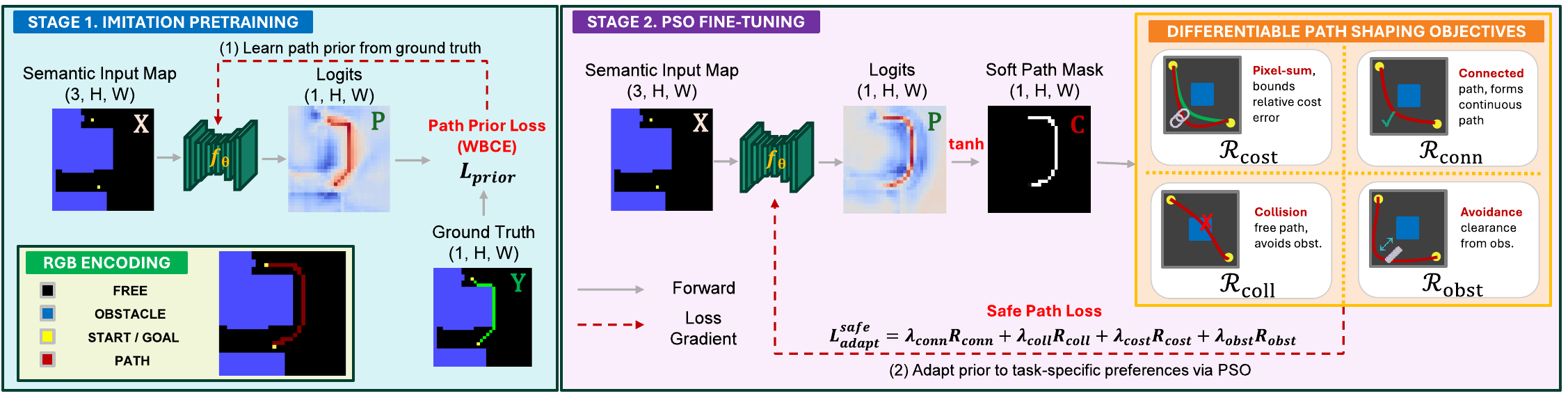}  
    \caption{%
    Overview of FlexPath.
    Stage~1 (left):
    Given an input map $\mathbf{X}$, a neural network $f_\theta$ predicts a raw path field $\mathbf{P}$ (logits), learning a path-connectivity prior via weighted binary cross-entropy against rasterized planner demonstrations $\mathbf{Y}$.\\
    Stage~2 (right):
    The prior is adapted via differentiable Path Shape Objectives (PSOs) encoding task-specific preferences (e.g. obstacle clearance).
    Stage~2 is only supervised by PSO functionals on the soft path mask $\mathbf{C}$, no expert demonstrations.
}
    \label{fig:method_overview}
\end{figure*}

\subsection{Problem Formulation and Overview}
\label{sec:formulation}

We formulate path planning as a dense prediction problem over a semantic grid. The input is a rasterized grid ${\mathbf{X} \in \mathbb{R}^{3 \times H \times W}}$. Each pixel here is encoded as an RGB triplet representing its semantic class, such as free space, obstacle, start, goal, and task-specific annotations.
Unlike classical binary or scalar cost maps, this representation allows us to encode task context directly into the input space, without additional channels. To formulate differentiable training objectives, we also define binary indicator maps ${\mathbf{O},\;\mathbf{S},\;\mathbf{G} \in \{0,1\}^{H \times W}}$, for obstacle, start, and goal locations (these are used solely to define PSOs).

Given $\mathbf{X}$, a neural network $f_\theta$ predicts a dense field over the grid:
$\mathbf{P} = f_\theta(\mathbf{X}) \in \mathbb{R}^{H \times W}$.
A shifted $\tanh$ activation then maps $\mathbf{P}$ to a bounded soft path mask:
$\mathbf{C} = \tfrac{1}{2}\bigl(\tanh(\mathbf{P}) + 1\bigr) \in [0,1]^{H \times W}$, where each entry $C_{ij}$ encodes the confidence that pixel $(i,j)$ lies on a feasible path. We choose a shifted tanh over a sigmoid activation, as it provides twice the maximum gradient around the origin ($0.5$ vs. $0.25$), leading to improved gradient flow during training \cite{pmlr-v9-glorot10a}. 
Together, $\mathbf{C}$ forms a path-probability map (PPM). This serves as a continuous relaxation of a binary path indicator, enabling gradient-based optimization of the path.
FlexPath's two-stage process is shown in \Cref{fig:method_overview}: We first learn path structure via imitation learning, then adapt it to task-specific criteria using differentiable PSOs.
During adaptation, PSO functionals $\mathcal{R}_{(\cdot)}$ act as task objectives on $\mathbf{C}$ whose values form training losses $\mathcal{L}_{(\cdot)}$, differentiable w.r.t. $\mathbf{C}$.
Supervision in Stage~1 is provided by $\mathbf{Y} \in \{0,1\}^{H \times W}$, which is the binary rasterization of an A*-optimal trajectory.

\subsection{Learning a Connectivity Prior}
\label{sec:feasibility}

The first stage learns to predict valid paths connecting start and end, thereby learning path structure. Demonstrations from a classical planner provide shortest-path solutions but primarily serve as demonstrations of spatial path-connectivity rather than an explicit objective to replicate.

\vspace{0.5em}

\noindent{\bfseries Imitation.} For each training environment, we compute a shortest path using A* and rasterize it as $\mathbf{Y} \in \{0,1\}^{H \times W}$.
The network is trained to predict $\mathbf{P}$, which are converted to pixel-wise probabilities as ${\hat{Y}_{ij} = \sigma(P_{ij})}$.

\vspace{0.5em}

\noindent{\bfseries Recall-biased objective.}
We bias the objective toward high recall to predict \emph{connected} traversable regions. 
In fact, this encourages a slightly over-complete prediction that preserves connectivity, which can be easily pruned during preference adaptation. This prior is intended only to provide a connected initialization for later adaptation, not to model the full set of feasible paths. The training minimizes a weighted binary cross-entropy loss,
\begin{equation}\label{eq:prior_loss}
\mathcal{L}_{\mathrm{prior}}
= -\sum_{i,j}\bigl[
      \alpha\,Y_{ij}\log\hat{Y}_{ij}
    + (1-\alpha)(1-Y_{ij})\log(1-\hat{Y}_{ij})
\bigr]
\end{equation}
where $\alpha>0.5$ increases the relative penalty on false negatives (missed path pixels), yielding a recall-oriented prior that better preserves connectivity for Stage~2 refinement. The effect of $\alpha$ is ablated in \Cref{sec:ablation}. Note that Stage~1 applies a standard sigmoid to $\mathbf{P}$, while Stage~2 uses the shifted $\tanh$.

\subsection{Preference Adaptation via Path Shape Objectives}
\label{sec:adaptation}

Given the connectivity prior, FlexPath adapts the soft path mask $\mathbf{C}$ to satisfy task-specific preferences without relearning path structure.
No label-based supervision is used; instead, optimization is driven by \emph{Path Shape Objectives (PSOs)}, (sub)differentiable functionals on $\mathbf{C}$.

We organize PSOs along two axes.
\emph{Structural PSOs} encode properties every valid path must satisfy: start-to-goal connectivity and collision avoidance.
\emph{Task PSOs} encode the preference.
We employ different connectivity formulations for optimal vs.\ non-optimal objectives to prevent conflicting gradient signals. Each PSO maps $\mathcal{R}_{(\cdot)} \colon [0,1]^{H \times W} \to \mathbb{R}$. The Stage~2 loss is a weighted combination of $\mathcal{R}_{(\cdot)}$ whose composition depends on the target objective.

\vspace{0.5em}

\noindent{\bfseries Structural PSOs.}
Structural PSOs strengthen task-agnostic geometric feasibility constraints, ensuring that predicted paths are collision-free and form a continuous connection between start and goal. While collision avoidance remains the same across tasks, the notion of connectivity varies by objective: obstacle-avoidance requires only reachability, whereas shortest-path tasks must also discourage inefficient routing. 

\paragraph{Suboptimal connectivity.}
For non-optimal objectives (e.g., clearance), only binary reachability is required.
A start marker $\mathbf{M}^{(0)} = \mathbf{S}$ is iteratively expanded:
$\mathbf{M}^{(t+1)} = \mathrm{MaxPool}_{3 \times 3}\!\bigl(\mathbf{M}^{(t)}\bigr) \odot \mathbf{C}$,
constraining propagation to the support of $\mathbf{C}$.
After $T$ iterations, we extract the value at the goal and apply a sharpened sigmoid:
$\mathcal{R}_{\mathrm{conn}}^{\mathrm{sub}} = -\sigma\!\bigl(\beta \cdot (M^{(T)}_{g_y, g_x} - 0.5)\bigr)$,
where $\beta > 0$ controls decision sharpness.

\paragraph{Optimal connectivity.}
For shortest-path objectives, the connectivity formulation penalizes inefficient routing.
Accumulated costs are propagated from the start location, with $M^{(0)}_{s_y,s_x} = 0$ and $10^6$ elsewhere.
At each iteration, for each pixel $(i,j)$ and its 8-connected neighbors $n_{kl} \in \mathcal{N}_{ij}$ with step cost $w_{k-i, l-j}$ ($\sqrt{2}$ diagonal, $1$ cardinal, $0.1$ center), a soft-minimum is computed via log-sum-exp:
\begin{equation}
    \tilde{M}^{(t)}_{ij} = -\frac{1}{\tau_t} \log \!\sum_{n_{kl} \in \mathcal{N}_{ij}} \exp\!\Bigl(-\tau_t \cdot (M^{(t-1)}_{kl} + w_{k-i,\,l-j})\Bigr)
    \label{eq:softmin}
\end{equation}
where $\tau_t$ is annealed temperature.
The cost map is updated:
\begin{equation}
    M^{(t)}_{ij} = \tilde{M}^{(t)}_{ij} + (1 - C_{ij}) + 10^6 \cdot O_{ij}
    \label{eq:cost_update}
\end{equation}
penalizing low-confidence regions and blocking obstacles.
After $T$ steps:
\begin{equation}
    \mathcal{R}_{\mathrm{conn}}^{\mathrm{opt}} = \min\!\bigl(M^{(T)}_{g_y, g_x},\; \mathcal{C}_{\max}\bigr)
    \label{eq:opt_conn}
\end{equation}
where $\mathcal{C}_{\max} = (\sqrt{2} + 1)\,T + 1$ clamps the cost to prevent uninformative gradients from the large initialization constant when the goal is unreachable (gradient will be zero).

\paragraph{Collision avoidance.} 
We compute the maximum path confidence on an obstacle: $\mathcal{R}_{\mathrm{coll}} = \max\bigl(\mathbf{C} \odot \mathbf{O}\bigr).$
Minimizing $\mathcal{R}_{\mathrm{coll}}$ penalizes the maximum path activation on obstacle cells, progressively reducing all such activations to zero.
\vspace{0.5em}

\noindent{\bfseries Task PSOs.}
While Structural PSOs strengthen geometric feasibility, Task PSOs encode the objective-dependent preferences that specify which path is desirable. These operate on the same confidence map $\mathbf{C}$ shaping its distribution to reflect task-specific criteria such as efficiency or obstacle clearance.

\paragraph{Cost minimization.}
Discrete path cost is approximated via pixel-wise convolution.
Let $\mathbf{K}_{\mathrm{cost}}$ encode step costs ($1$ cardinal, $\sqrt{2}$ diagonal) and $\mathbf{K}_{\mathrm{avg}}$ be a uniform $3{\times}3$ box filter.
The pixel-wise cost approximation is:
\begin{equation}
    \mathbf{C}_{\mathrm{cost}} = \mathbf{C} \odot \frac{\mathbf{C} \ast \mathbf{K}_{\mathrm{cost}}}{\mathbf{C} \ast \mathbf{K}_{\mathrm{avg}} + \varepsilon}
    \label{eq:cost_approx}
\end{equation}
where $\ast$ denotes convolution.
The PSO is then defined as the normalized sum of pixel-wise costs:
\begin{equation}
    \mathcal{R}_{\mathrm{cost}} =
    \frac{\sum_{i,j}\mathbf{C}_{\mathrm{cost}}(i,j)}
    {H \times W \times \sqrt{2}}
    \label{eq:cost_pso}
\end{equation}

\paragraph{Obstacle clearance.}
A differentiable penalty is defined for paths that approach closer than $d_{\min}$ to any obstacle.
For each pixel $(i,j)$, we compute a soft approximation of the distance to the nearest obstacle via windowed log-sum-exp:
\begin{equation}
    d^*_{ij} = -\frac{1}{\tau}\log \sum_{k,\ell \;\in\; \mathcal{W}(i,j)} 
    \exp\!\bigl(-\tau \cdot \tilde{d}_{ijk\ell}\bigr)
    \label{eq:soft_min_dist}
\end{equation}
where $\mathcal{W}(i,j) = [i \pm d_{\min}]_H \times [j \pm d_{\min}]_W$ 
denotes the $d_{\min}$-relevant neighborhood window around $(i,j)$ and $\tilde{d}_{ijk\ell}$ the pairwise distance from pixel $(i,j)$ to pixel $(k,\ell)$. Distances to non-obstacle cells $(k,\ell)$ are masked with a large constant ($10^6$), and $\tau > 0$ controls approximation sharpness.
A proximity penalty, which is nonzero only when $d^*_{ij} < d_{\min}$, is computed and masked by path confidence:
\begin{equation}
    \rho_{ij} = C_{ij} \cdot \frac{\max\bigl(0,\; d_{\min} - d^*_{ij}\bigr)}{d_{\min}}
    \label{eq:proximity}
\end{equation}
The clearance PSO combines the mean and maximum of the penalties to provide a balanced gradient signal, such that $\mathcal{R}_{\mathrm{obst}}
    =
    \frac{1}{2}
    \left(
        \operatorname{mean}\rho
        +
        \max\rho
    \right).$

\vspace{0.5em}
\noindent{\bfseries Combined Training Objective.} Our Stage~2 loss guides towards different navigation behaviors depending on the preference by integrating shared structural constraints with task-specific PSOs. Each PSO aggregation is a weighted linear combination of its respective objective components with scalars $\lambda_{(\cdot)}$. We define two general aggregations, one for optimal (shortest-path) pathfinding and one for sub-optimal objectives (like obstacle clearance) that require detours from the shortest-possible path. Both variants include collision term $\mathcal{R}_{\mathrm{coll}}$ and spatial-cost term $\mathcal{R}_{\mathrm{cost}}$, while the connectivity term is task-specific: $\mathcal{R}_{\mathrm{conn}}^{\mathrm{opt}}$ promotes efficient routing, whereas $\mathcal{R}_{\mathrm{conn}}^{\mathrm{sub}}$ permits detours required for clearance. The safe-path objective additionally includes $\mathcal{R}_{\mathrm{obst}}$.
In both cases, connectivity is computed bidirectionally and then averaged. We tuned loss weights $\lambda_{(\cdot)}$ separately for each objective and provide them in the supplementary material.

\vspace{0.5em}
\noindent{\bfseries Inference.} At test time, a graph-based planner is used to extract a discrete trajectory, where we convert the predicted soft path mask into an inadmissible secondary heuristic by inversion and scaling,
\(h_C(v)=\gamma(1-C_v)\), so high-confidence path cells receive low heuristic cost value, while the admissible octile heuristic remains the anchor for search. The learned heuristic therefore biases expansion toward the adapted objective, whereas validity is enforced by graph search on the collision-free grid. For shortest-path objectives, we use Focal Search~\cite{FocalSearch}, and for all other objectives, Multi-Heuristic A*~\cite{MultiHeuristicAstar}. Algorithmic details and pseudocode are provided in the supplementary.

\section{Experiments}
\label{sec:experiments}

We evaluate FlexPath in two settings.
First, we assess shortest-path planning performance against established baselines.
Second, we show that a single pretrained model can be flexibly adapted to a range of non-shortest-path objectives.

\noindent{\bfseries Datasets.}
\label{sec:datasets}
We adopt a cross-dataset protocol: the model is trained on a single source dataset
and evaluated without retraining on all others.
\textbf{Tiled Motion Planning (TMP)}~\cite{yonetani2021pathplanningusingneural,
bhardwaj2017learningheuristicsearchimitation} constructs grid environments (512k/64k/64k split) by composing
canonical tile patterns; we use TransPath's 640k-environment extension at $64{\times}64$
resolution \cite{kirilenko2022transpathlearningheuristicsgridbased} as primary
training source and an in-distribution test bed because it is the standard benchmark used by prior learned path-guidance methods, enabling direct comparison under a controlled setting.
Three datasets serve as zero-shot evaluation benchmarks:
\textbf{VoxelGym (VG)} \cite{Babu_2023}, $64{\times}64$ grids derived from AirSim-NH~\cite{airsim2017fsr} (50k maps, 40k/10k split);
\textbf{City/Street Map (CSM)} \cite{6194296}, real-world city maps; and
\textbf{Starcraft (SC)} \cite{6194296}, containing diverse maps from the game Starcraft. We chose these to assess our method on a highly diverse set of topologies that significantly differ from TMP. 
We additionally report results with VoxelGym as the training source in the supplementary material.

\noindent{\bfseries Evaluation Metrics.}
\label{sec:metrics}
We report four shortest-path metrics: three adopted from TransPath~\cite{kirilenko2022transpathlearningheuristicsgridbased} and one new that measures reliance on the downstream planner.
\begin{itemize}
    \item \textbf{Cost Factor}: ratio of planned path cost to
    A*-optimal cost; $1.0$ is exact recovery~\cite{kirilenko2022transpathlearningheuristicsgridbased}.
    \item \textbf{Expansion Ratio}: nodes expanded on the
    learned map relative to A* on the full grid~\cite{kirilenko2022transpathlearningheuristicsgridbased}.
    \item \textbf{Optimal Found Ratio}: fraction of paths
    achieving optimal cost~\cite{VlastelicaEtal2020:BBoxSolvers}.
    \item \textbf{Hard Validity}: fraction of valid paths
    when restricting search to cells with $\geq$50\% confidence. Reported only for methods producing explicit path-probability maps
\end{itemize}
Cost Factor and Optimal Found Ratio assess the path cost, Expansion Ratio measures efficiency and Hard Validity measures how close the prediction itself is to a connected path.

\noindent{\bfseries Implementation Details.}
\label{sec:implementation}
We adopt a U-Net~\cite{UNET} with three ResNet blocks~\cite{He2015DeepRL} per encoder/decoder stage (base channels: 64), nearest-neighbor down/upsampling, and four Transformer blocks~\cite{Vaswani2017AttentionIA} (four heads, learned positional encoding) at the bottleneck. 
We train in PyTorch 2.8~\cite{Ansel_PyTorch_2_Faster_2024} with AdamW~\cite{Loshchilov2017DecoupledWD} and cosine decay~\cite{Loshchilov2016SGDRSG}. Stage~1 trains for 250 epochs (lr $2.5{\times}10^{-5}$, batch 512); Stage~2 for 250 epochs (lr $1.0{\times}10^{-5}$, batch 128) to ensure PSO convergence. On one A100 80GB, Stage~1 takes $\sim$10h, Stage~2 $\sim$7h per objective with a shared prior; by comparison, TransPath requires $\sim$4h for full training, and Neural~A*, iA* and DAA* $\sim$60h. To assess sensitivity to the training budget, we additionally train FlexPath using a reduced budget of 2h for Stage~1 and 1h for Stage~2. Full hyperparameters are given in the supplementary.

\subsection{Shortest-Path Results}
\label{sec:shortest_path_results}

We compare against Neural A* \cite{yonetani2021pathplanningusingneural}, iA* \cite{iA*}, DAA* \cite{DAA*}, and TransPath \cite{kirilenko2022transpathlearningheuristicsgridbased} on TMP 640k, and ablate our two-stage training (PT, FT, PT+FT; see \Cref{tab:tmp_results}). We evaluate each baseline with its intended planning backend, following standard practice for learned-search methods. Neural A*, iA*, and DAA* are retrained with the same backbone to control for model capacity, while TransPath uses the official checkpoint and search algorithm (Focal Search). FlexPath uses the same Focal Search setup as TransPath, with the same weight ($w=2$). All metrics are reported from one fixed-seed training run per method, consistent with common practice in learned grid-based path-planning benchmarks; reported variance is computed over evaluation instances.

Note that Neural A*, iA*, DAA*, and TransPath receive binary occupancy inputs; FlexPath uses the same binary occupancy as input, but re-encodes it into RGB-space via fixed, deterministic mappings, thus introducing no additional environmental information.

\begin{table}[t]
\centering
{\small
\setlength{\tabcolsep}{1.8pt}
\begin{tabular*}{\columnwidth}{@{\extracolsep{\fill}} llcccc}
\toprule
\small{Data} & \small{Method}
& Cost  $\downarrow$
& Exp.\ $\downarrow$
& H.\ Val.  $\uparrow$
& Opt.\ F. $\uparrow$ \\
\midrule
\multirow{8}{*}{\rotatebox[origin=c]{90}{\small TMP}}
& Neural A* 
  & $1.007\!\pm\!.02$ & $0.318\!\pm\!.20$ & -- & $0.702$ \\
& iA* 
  & $1.041\!\pm\!.05$ & $0.367\!\pm\!.21$ & -- & $0.227$ \\
& DAA*  
  & $1.007{\pm}.01$ & $0.231{\pm}.21$ & -- & $0.617$ \\
& TransPath 
  & $1.005\!\pm\!.02$ & $0.189\!\pm\!.16$ & $0.892$ & $0.750$ \\
& Ours PT
  & $1.010\!\pm\!.01$ & $0.167\!\pm\!.15$ & $0.984$ & $0.470$ \\
& Ours FT
  & $1.004\!\pm\!.01$ & $0.168\!\pm\!.11$ & $0.986$ & $0.795$ \\
& Ours (Eff.)
  & $1.004{\pm}.01$ & $0.170{\pm}.11$ & $\mathbf{0.992}$ & 0.818 \\
& Ours 
  & $\mathbf{1.002\!\pm\!.01}$ & $\mathbf{0.163\!\pm\!.11}$ & $0.991$ & $\mathbf{0.886}$ \\
\midrule
\multirow{6}{*}{\rotatebox[origin=c]{90}{\small VG}}
& Neural A*  & $1.004\!\pm\!.01$ & $0.305\!\pm\!.15$ & -- & $0.822$ \\
& iA*  & $1.020\!\pm\!.03$ & $0.304\!\pm\!.16$ & -- & $0.543$ \\
& DAA*        & $1.004{\pm}.01$ & $0.263{\pm}.18$ & -- & $0.810$ \\
& TransPath  & $\mathbf{1.002\!\pm\!.01}$ & $\mathbf{0.245\!\pm\!.13}$ & $0.976$ & $0.901$ \\
& Ours (Eff.) & $1.002{\pm}.01$ & $0.261{\pm}.13$ & $\mathbf{0.999}$ & 0.950 \\
& Ours & $1.002\!\pm\!.02$ & $0.253\!\pm\!.13$ & $0.998$ & $\mathbf{0.956}$ \\
\midrule
\multirow{6}{*}{\rotatebox[origin=c]{90}{\small CSM}}
& Neural A*  & $1.016\!\pm\!.03$ & $0.324\!\pm\!.18$ & -- & $0.502$ \\
& iA*  & $1.038\!\pm\!.04$ & $0.295\!\pm\!.16$ & -- & $0.210$ \\
& DAA*        & $1.019{\pm}.04$ & $0.293{\pm}.24$ & -- & $0.439$ \\
& TransPath  & $1.029\!\pm\!.06$ & $0.430\!\pm\!.49$ & $0.333$ & $0.348$ \\
& Ours (Eff.) & $\mathbf{1.012{\pm}.04}$ & $0.303{\pm}.19$ & $\mathbf{0.902}$ & 0.632 \\
& Ours & $1.013\!\pm\!.04$ & $\mathbf{0.281\!\pm\!.17}$ & $0.895$ & $\mathbf{0.651}$ \\
\midrule
\multirow{6}{*}{\rotatebox[origin=c]{90}{\small SC}}
& Neural A*  & $1.020\!\pm\!.03$ & $0.430\!\pm\!.23$ & -- & $0.373$ \\
& iA*  & $1.038\!\pm\!.04$ & $0.399\!\pm\!.22$ & -- & $0.211$ \\
& DAA*        & $1.032{\pm}.04$ & $\mathbf{0.270\!\pm\!.23}$ & -- & $0.257$ \\
& TransPath  & $1.048\!\pm\!.08$ & $0.490\!\pm\!.46$ & $0
.164$ & $0.198$ \\
& Ours (Eff.) & $\mathbf{1.015{\pm}.04}$ & $0.296{\pm}.21$ & $\mathbf{0.875}$ & 0.586 \\
& Ours  & $1.019\!\pm\!.04$ & $0.272\!\pm\!.19$ & $0.840$ & $\mathbf{0.532}$ \\
\bottomrule
\end{tabular*}}
\caption{Shortest-path and zero-shot generalization results.
All models trained on TMP~640k. Ours refers to the full pipeline, Eff. to the reduced training cost run. PT: Stage~1 only, FT: Stage~2 only.}
\label{tab:tmp_results}
\end{table}

\begin{figure}[t]
    \centering
    \includegraphics[width=\columnwidth]{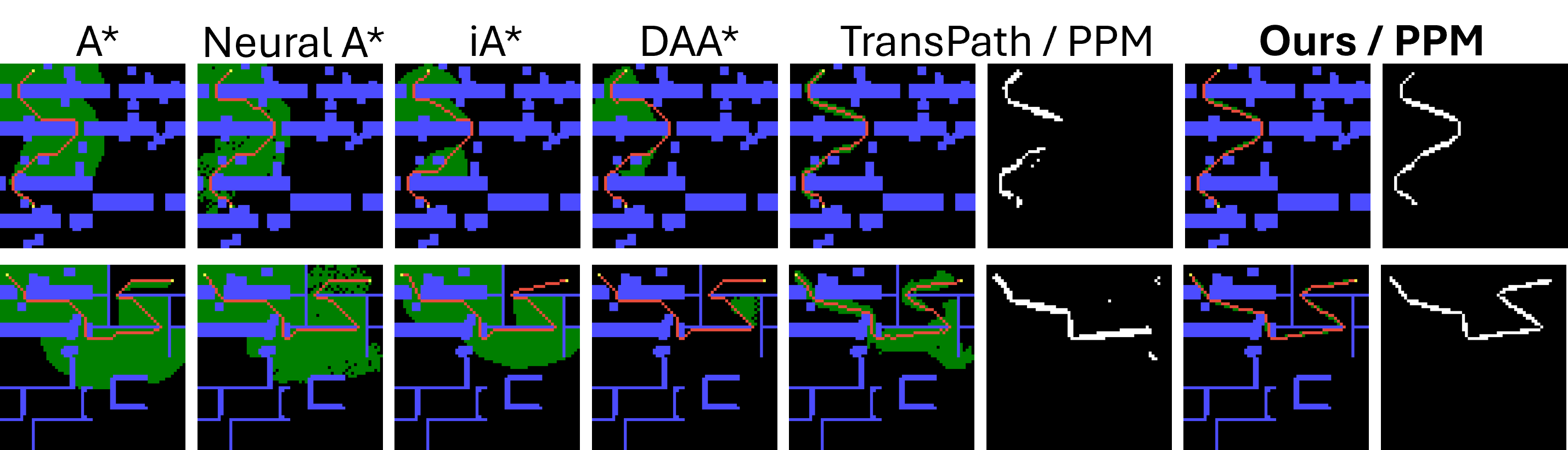}
    \caption{%
        Qualitative comparison on TMP~640k.
        Selected examples of the pathfinding results. Expanded nodes are shown in green, and the resulting path is shown in red.
        PPMs are only shown for methods that explicitly produce them.
    }
    \label{fig:qualitative_tmp}
\end{figure}

\Cref{tab:tmp_results} reports shortest-path performance on TMP~640k. PT and FT denote models trained using only Stage~1 and Stage~2 from a random initialization, respectively; Eff. denotes the reduced-budget 2h/1h training run. Overall, FlexPath finds an optimal solution for $88.6\%$ of all scenarios, $13.6$ percentage points more often than second-best TransPath, and reduces avg. path cost to only $0.2\%$ more than A* ($0.5\% \to 0.2\%)$. At the same time FlexPath lowers avg. search effort by $13.8\%$ ($0.189 \to 0.163$). Notably, both cost and effort improve jointly, ruling out a simple unbalanced optimality/efficiency tradeoff. Hard Validity further supports this: FlexPath's raw PPM already yields a valid path $99.1\%$ of the time (vs. $89.2\%$ for TransPath), indicating high quality path representations. \Cref{fig:qualitative_tmp} illustrates why; FlexPath's PPM concentrates tightly around the optimal corridor, whereas TransPath's is more diffuse.

FlexPath also generalizes well to unseen environments. Across VG, CSM, and SC (\Cref{tab:tmp_results}, bottom), our method consistently ranks first or second in Cost Factor and Expansion Ratio, and achieves the top Hard Validity and Optimal Found Ratio on all three benchmarks, indicating robust cross-benchmark performance.
The advantage is most pronounced on CSM and SC, where baseline validity and efficiency degrade substantially under distribution shift. The reduced-budget run achieves performance close to the full training schedule despite using substantially less optimization time, indicating that the proposed training procedure reaches strong performance before full convergence. Full per-dataset qualitative results are provided in the supplementary.

\subsection{Inference and Planning Time Comparison}

We first compare against TransPath using Focal Search (matching their paper's setup). At 64x64 resolution, TransPath's planning takes an avg. of $0.71$ ms compared to $0.47$ for FlexPath and $4.49$ ms for naive Focal Search. Neural network inference time is 6.5ms/787k parameters for FlexPath versus TransPath’s 6.6ms/944k. To evaluate scalability, we retrained our model at 256×256 resolution and compared total time (inference + planning) against naïve Focal Search. FlexPath's total time (inference+planning) increases by only a factor of ~4.6 ($7.05 \to 32.65$), whereas naive Focal Search increases by a factor of ~27.8 ($4.49 \to 124.87$).
In a second comparison against Neural A*, iA*, and DAA*, we use their native Neural A* search backend across all models to ensure a fair, implementation-independent evaluation. Under this shared backend, FlexPath achieves the fastest avg. planning time at $38$ ms, compared to $66$ (DAA*), $77$ (Neural A*), and $92$ ms (iA*). Since all models share the same neural network architecture, avg. inference time is $6.5$ ms in all cases.

\subsection{Ablation Studies}
\label{sec:ablation}

We ablate key design choices on TMP~640k using the Stage~1 (PT) model (\Cref{tab:ablation}). Each row modifies one component relative to the baseline configuration (Weighted BCE with
$\alpha{=}0.95$, RGB, UNet+Att). The two-stage training paradigm is additionally ablated in \Cref{tab:tmp_results}.

\vspace{0.5em}
\noindent{\bfseries Two-stage training.}
The training paradigm is ablated in \cref{tab:tmp_results}. PT alone learns a strong connectivity prior (Hard Validity $0.984$) but a lower Optimal Found Ratio ($0.470$). FT alone reaches competitive validity but plateaus at $0.795$ Optimal Found Ratio and, most crucially,
collapses for non-optimal objectives (see \Cref{sec:flexibility}). The combined pipeline (PT+FT/Eff.) achieves the best results across all metrics, confirming that the two stages are complementary.

\vspace{0.5em}
\noindent{\bfseries Loss formulation.}
Both MSE and unweighted BCE ($\alpha{=}0.5$) severely degrade
Hard Validity ($0.089$ and $0.080$), producing fragmented, less connected, predictions. This validates our recall-biased
weighting in \cref{eq:prior_loss}: penalizing false negatives
($\alpha{>}0.5$) yields spatially connected priors that remain intact
under hard thresholding and stabilize Stage~2 refinement.

\vspace{0.5em}
\noindent{\bfseries Input representation.}
Binary occupancy produces similar hard validity and efficiency results on TMP~640k, but RGB is retained to preserve the flexibility of adding new semantic classes without requiring the model to relearn geometric path feasibility. However, fine-tuning the binary occupancy model fully matches the RGB model, ruling out that shortest-path gains come from RGB (metrics in supplementary).


\begin{table}[t]
\centering
{\small
\setlength{\tabcolsep}{1.8pt}
\begin{tabular*}{\columnwidth}{@{\extracolsep{\fill}} lcccc}
\toprule
Variant
& Cost $\downarrow$
& Exp.\ $\downarrow$
& H. Val. $\uparrow$
& Opt.\ F. $\uparrow$ \\
\midrule
Ours (PT)           & $1.010\!\pm\!.01$ & $0.167\!\pm\!.15$ & $0.984$ & $0.470$ \\
\midrule
\rowcolor[HTML]{E6E6E6}
\multicolumn{5}{c}{\textit{Loss formulation}} \\
MSE                             & $1.010\!\pm\!.02$ & $0.163\!\pm\!.14$ & $0.089$ & $0.509$ \\
BCE                             & $1.010\!\pm\!.01$ & $0.160\!\pm\!.13$ & $0.080$ & $0.499$ \\
\midrule
\rowcolor[HTML]{E6E6E6}
\multicolumn{5}{c}{\textit{Input representation}} \\
Bin. occ.                    & $1.012\!\pm\!.02$ & $0.174\!\pm\!.15$ & $0.986$ & $0.386$ \\
\bottomrule
\end{tabular*}}
\caption{Ablation studies on TMP~640k.
Each row modifies one component relative to the Stage~1 baseline
(Weighted BCE, RGB). 
}
\label{tab:ablation}
\end{table}

\section{Flexibility Across Objectives}
\label{sec:flexibility}

A key feature of FlexPath is that a single pretrained connectivity-prior can be reused across planning objectives, requiring only Stage~2 adaptation rather than relearning path structure. Most crucially, adapting from the connectivity-prior instead of a random initialization improves final performance or avoids full model collapse. We demonstrate this by adapting the same Stage~1 model to tasks that require qualitatively different routing behaviors, namely obstacle clearance, class-conditioned obstacle clearance and waypoint following. We evaluate the performance against a cost-map shaped (details in supplementary) classical WA* (baseline) and compare against adapting from a random initialization instead of the connectivity-prior (FT). We report:
\begin{itemize}
    \item \textbf{Expansion Ratio}: defined in~\cref{sec:metrics}.
    \item \textbf{Clearance Satisfaction}: fraction of feasible scenarios whose paths fully satisfy the clearance threshold (clearance objectives only).
    \item \textbf{Waypoint Satisfaction}: ratio of start-to-waypoint-to-goal connected paths when restricting search to cells with $\geq$50\% confidence (waypoint objective only).
\end{itemize}

\vspace{0.5em}
\noindent{\bfseries Obstacle Clearance.}
\label{sec:obstacle_clearance}
We first adapt the pretrained prior to favor safety over efficiency by preferring a soft minimum distance $d_{\min}$ from obstacles, using the sub-optimal PSO aggregation specified in \cref{sec:adaptation}. For example, \(d_{\min}=2\) desires at least one free pixel between the path and the obstacles. We compute the clearance satisfaction only on environments where this clearance is feasible ($54.45\%$). 

\begin{figure}[b]
    \centering
    \includegraphics[width=1\linewidth]{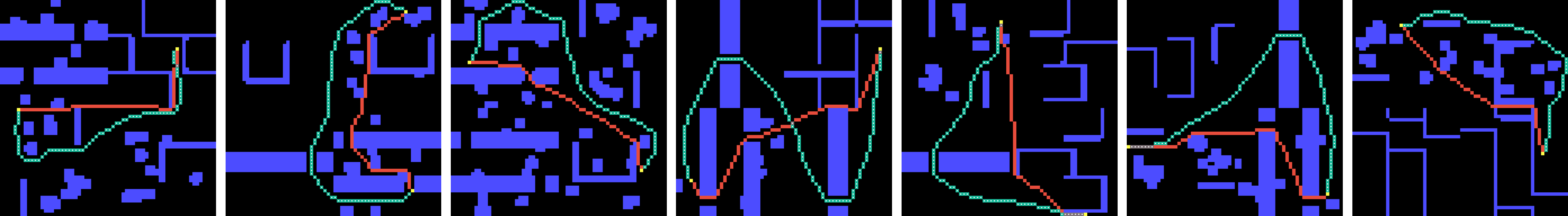}
    \caption{%
        The same pretrained model adapted to two objectives: clearance-constrained planning (aquamarine/dotted, $d_{\min}=2$) and shortest-path (red).
    }
    \label{fig:qualitative_clearance}
\end{figure}

\Cref{tab:adaptation_results} shows the necessity of the pretraining. The model adapted from a random initialization (FT) suffers from all-zero collapse and consequently achieves $0.0\%$ clearance satisfaction and high expansion ratio. Our full pipeline achieves the most favorable results, improving clearance satisfaction compared to the classical cost-map shaping WA* baseline ($0.836 \to 0.962$) while keeping expansion ratio low ($0.16$). Qualitative results (\Cref{fig:qualitative_clearance}) also confirm that our adapted model shifts paths toward wider corridors. 
We provide a more detailed evaluation including feasible vs. infeasible environment analysis, results for \(d_{\min}=3\) and cross-distribution performance in the supplementary.

\begin{table}[t]
\centering
{\small
\setlength{\tabcolsep}{2.8pt}
\begin{tabular*}{\columnwidth}{@{\extracolsep{\fill}}llcc}
\toprule
Objective & Method & Sat.\ $\uparrow$ & Exp.\ $\downarrow$ \\
\midrule

\multirow{2}{*}{\shortstack[l]{Clearance\\($d_{\min}=2$)}}
& Baseline / FT
& $0.836$ / $0.0$
& $0.41\!\pm\!.30$ / $4.56\!\pm\!2.80$ \\
& PT+FT
& $\mathbf{0.962}$
& $\mathbf{0.16\!\pm\!.23}$ \\
\midrule

\multirow{2}{*}{\shortstack[l]{Class-cond.\\clearance}}
& Baseline / FT
& $0.660$ / $0.0$
& $0.51\!\pm\!.47$ / $4.56\!\pm\!2.80$ \\
& PT+FT
& $\mathbf{0.684}$
& $\mathbf{0.19\!\pm\!.29}$ \\
\midrule

\multirow{2}{*}{\shortstack[l]{Waypoint\\following}}
& Baseline / FT
& $\mathbf{1.0}$ / 0.973
& $1.15\!\pm\!1.00$ /  $0.20\!\pm\!0.15$ \\
& PT+FT
& $0.984$
& $\mathbf{0.19\!\pm\!.14}$ \\

\bottomrule
\end{tabular*}
}
\caption{Preference adaptation on TMP~640k. Sat.\ denotes full satisfaction of the requested preference:
clearance and class-conditioned clearance are computed on feasible instances,
while waypoint Sat.\ is start-to-waypoint-to-goal success. The first value in each Baseline / FT cell is the classical baseline;
the second is Stage~2 training from random initialization. PT+FT fine-tunes the shared connectivity prior.}
\label{tab:adaptation_results}
\end{table}

\vspace{0.5em}
\noindent{\bfseries Class-conditioned clearance.}
We introduce a second obstacle class with a larger minimum clearance ($d_{\min}=4$ vs. $2$ for standard obstacles), which enables danger-aware avoidance. This class is randomly added to free cells of TMP samples. The task loss is a simple modification of the sub-optimal PSO aggregation specified in \cref{sec:adaptation}, where $\mathcal{R}_{\mathrm{obst}}$ is computed for each obstacle class and its desired clearance separately and then averaged. Across 64k test environments, our model fully maintains the desired clearance to those obstacles in 68.4\% of possible cases, beating the classical baseline ($0.660 \to 0.684$), while preserving strong avoidance of standard obstacles. Again, fine-tuning from a random initialization instead of the connectivity-prior leads to full collapse. Full metrics and the detailed dataset generation process are provided in the supplementary.

\begin{figure}[t]
    \centering
    \includegraphics[width=1\linewidth]{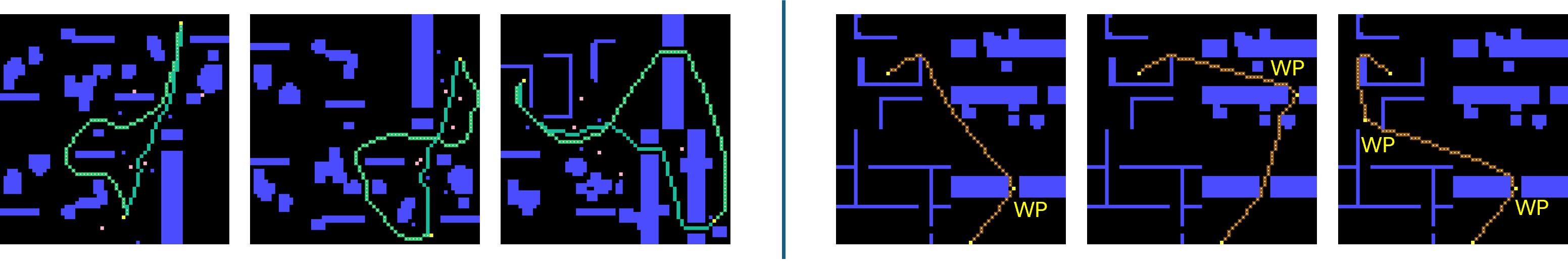}
    \caption{%
        Class-conditioned clearance and waypoint guidance. 
        Left: Trajectory from the multi-clearance-class model (green/dotted). 
        Pink regions encourage a larger minimum clearance of
        $d_{\min}=4$ pixels, compared to standard obstacle avoidance (aquamarine) with $d_{\min}=2$.
        Right: Waypoint-guided planning (brown/dotted).
        Additional waypoint pixels (yellow/WP) should be traversed, if present.
    }
    \label{fig:semanticFigure}
\end{figure}

\vspace{0.5em}
\noindent{\bfseries Waypoint following.}
Intermediate waypoints act as soft constraints, guiding the trajectory through specified locations. The task loss is a simple modification of the optimal PSO aggregation specified in \cref{sec:adaptation}, where $\mathcal{R}^{opt}_{\mathrm{conn}}$ is computed from start-to-waypoint and waypoint-to-target and then averaged. Our adapted model successfully connects start to waypoint to goal in 98.4\% of all test cases. Adapting from a random initialization instead of the prior leads to worse results ($0.973$). Classical search can solve this problem perfectly by applying A* from start-to-waypoint and waypoint-to-goal separately, but at higher search cost. \Cref{fig:semanticFigure} shows visualizations for both objectives. Detailed analysis, including bidirectional adaptation experiments, is provided in the supplementary.

\section{Conclusion}
\label{sec:conclusion}

FlexPath shows that separating reusable path-structure learning from objective-specific preference adaptation is an effective design principle for learned grid planning. Learning a recall-oriented connected-route prior and subsequently refining it with differentiable Path Shape Objectives improves standard shortest-path planning itself: the two stages are complementary, jointly yielding stronger path quality and search efficiency than either component alone. The resulting search guidance also transfers robustly across unseen map domains.

Beyond shortest-path planning, the same Stage-1 initialization can be adapted to clearance, class-conditioned avoidance, and waypoint guidance without collecting additional pixel-level path labels or retraining the full model from scratch. This demonstrates that learned search guidance need not remain bound to the objective represented by its demonstrations, but can be refined toward alternative path preferences through objective-level supervision.

Current limitations include the focus on 2D rasterized environments and manually specified PSO forms and loss weights. Future work could extend the framework to 3D and continuous planning domains, investigate more parameter-efficient or task-conditioned adaptation, and learn objective trade-offs directly from data or user preferences.

\bibliography{aaai2027}


\clearpage
\appendix

\setcounter{figure}{0}
\setcounter{table}{0}
\setcounter{equation}{0}
\setcounter{algorithm}{0}

\renewcommand{\thefigure}{S\arabic{figure}}
\renewcommand{\thetable}{S\arabic{table}}
\renewcommand{\theequation}{S\arabic{equation}}
\renewcommand{\thealgorithm}{S\arabic{algorithm}}


\noindent{\bfseries Overview.}
This supplementary material provides details omitted from the main paper due to space constraints.
\Cref{sec:impl} describes implementation specifics: input encoding, loss weights, dataset modification, inference, architecture, and hyperparameters used in the training.
\Cref{sec:ext-clear}, \Cref{sec:ext-adapt} and \Cref{sec:ext-sp} add additional quantitative and qualitative comparisons for the shortest-path, clearance, and adaptation experiments, including classical cost-map baselines and a bidirectional analysis.
\Cref{sec:ablations} reports further ablations on PSO weight sensitivity and on training instability when Stage~1 pretraining is skipped.
\Cref{sec:qual} collects additional visualizations.
All notation follows the main paper.



\section{Additional Implementation Details}
\label{sec:impl}

\subsection{Color Conventions}
\begin{figure}
    \centering
    \includegraphics[width=\columnwidth]{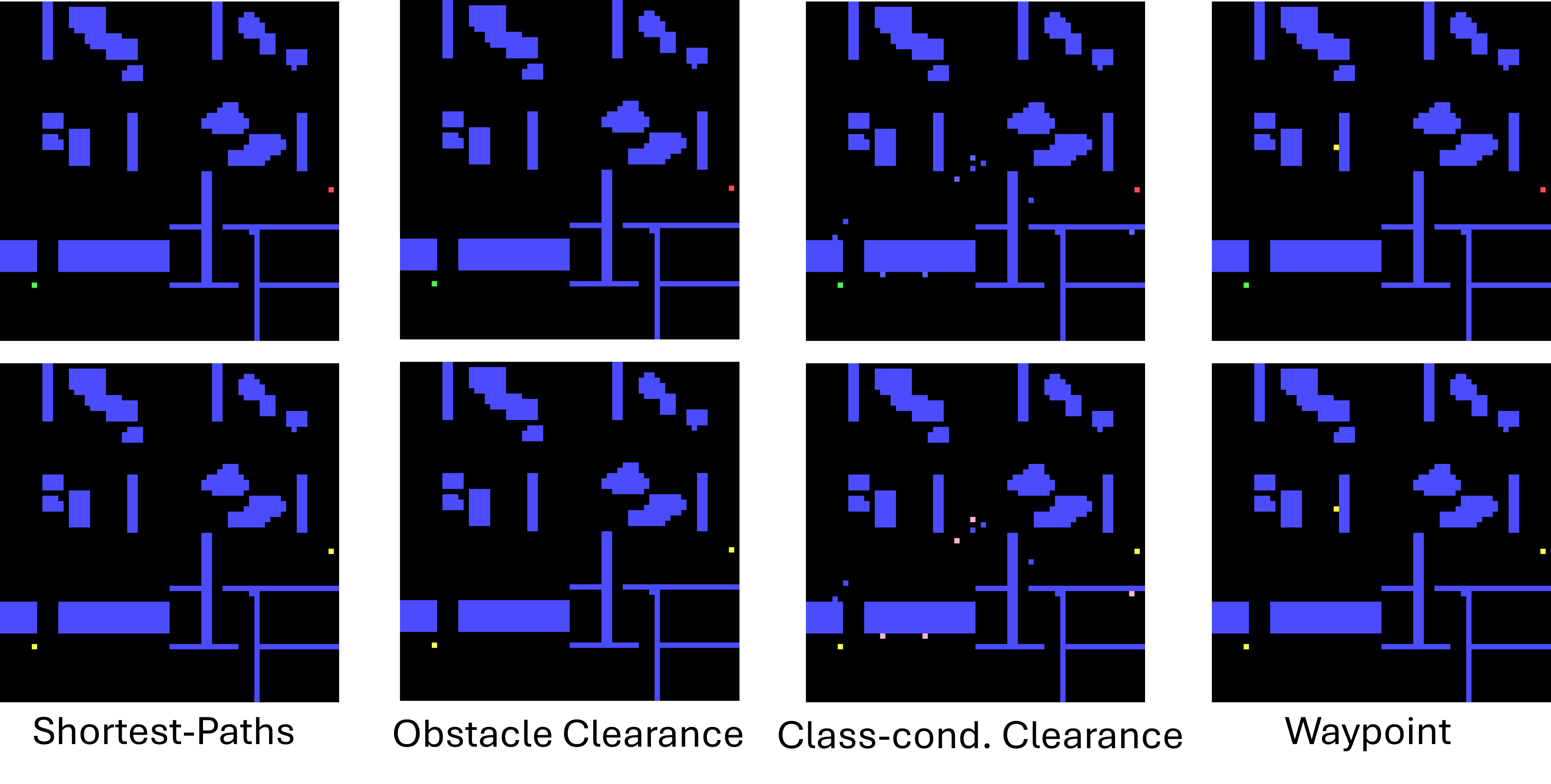}
    \caption{
    Representative RGB encodings for each objective. The upper row shows the input to the model, the bottom one shows the remapped color encodings used for better visualization throughout this paper.}
    \label{fig:task-grids}
\end{figure}

The RGB values used during training differ from those shown in our result
figures: we remap colors in all visualizations only for better readability.
\Cref{tab:colors} denotes the exact training-time RGB encoding for
each semantic class across all task settings, and \Cref{fig:task-grids} shows a representative input grid for each task.

\begin{table*}[t]
  \centering
  \begin{tabular}{l c c c c c c}
    \toprule
    & \textbf{Free space}
    & \textbf{Obstacle}
    & \textbf{Start}
    & \textbf{Goal}
    & \textbf{Sem. Obstacle}
    & \textbf{Waypoint} \\
    \midrule
    RGB
      & (0,0,0) & (76,76,255) & (255,76,76)
      & (76,255,76) & (100,100,255) & (255,255,76) \\[4pt]
    Color
      & \textcolor{clrFree}{\rule{8pt}{8pt}}
      & \textcolor{clrObst}{\rule{8pt}{8pt}}
      & \textcolor{clrStart}{\rule{8pt}{8pt}}
      & \textcolor{clrGoal}{\rule{8pt}{8pt}}
      & \textcolor{clrSemanticObst}{\rule{8pt}{8pt}} 
      & \textcolor{clrWaypoint}{\rule{8pt}{8pt}} \\
    \bottomrule
  \end{tabular}
    \caption{%
    Training-time RGB encoding for each semantic class.
    Visualization figures use remapped colors for readability
    (see ~\Cref{fig:task-grids}). }
  \label{tab:colors}
\end{table*}

\subsection{Loss Weights per Objective}
As each task prefers different planning criteria, the loss composition varies. Here we give the full loss formulations for class-conditioned clearance and waypoint following, which are omitted from the main paper. Both closely resemble the formulations for shortest paths or obstacle avoidance, differing only in one component.

\paragraph{Class-conditioned clearance loss.}
Obstacle clearance with reachability-only connectivity:
\begin{equation}
    \mathcal{L}_{\mathrm{adapt}}^{\mathrm{safe}} =
      \lambda_{\mathrm{conn}}\mathcal{R}_{\mathrm{conn}}^{\mathrm{sub}}
    + \lambda_{\mathrm{coll}}\mathcal{R}_{\mathrm{coll}}
    + \lambda_{\mathrm{obst}}\mathcal{R}^*_{\mathrm{obst}}
    + \lambda_{\mathrm{cost}}\mathcal{R}_{\mathrm{cost}}
    \label{eq:loss_clearance}
\end{equation}
where $\mathcal{R}^*_{\mathrm{obst}}$ is the average of $\mathcal{R}_{\mathrm{obst}}$ applied to the second obstacle class with $d_{\min}=4$ and normal obstacles with $d_{\min}=2$.

\paragraph{Waypoint following loss.}
Cost minimization with cost-aware connectivity:
\begin{equation}
    \mathcal{L}_{\mathrm{adapt}}^{\mathrm{opt}} =
      \lambda_{\mathrm{conn}} \, \mathcal{R}_{\mathrm{conn}}^{\mathrm{opt}*}
    + \lambda_{\mathrm{coll}} \, \mathcal{R}_{\mathrm{coll}}
    + \lambda_{\mathrm{cost}} \, \mathcal{R}_{\mathrm{cost}}
    \label{eq:loss_optimal}
\end{equation}
where $\mathcal{R}_{\mathrm{conn}}^{\mathrm{opt}*}$ averages $\mathcal{R}_{\mathrm{conn}}^{\mathrm{opt}}$ computed from start to waypoint and waypoint to goal.
All weight values and annealing schedules are listed in \Cref{tab:hyperparams}.

\subsection{Dataset Modifications per Objective}
We applied modifications to the dataset for waypoint-following and class-conditioned clearance objectives to accommodate the additional required semantic classes. We keep the same base maps and start/goal configurations of TMP for both objectives (see \Cref{fig:task-grids} for a sample map).

\paragraph{Class-conditioned clearance.} We first construct a candidate space consisting of free, non-path cells that lie within a distance of at most 10 (octile distance) from the closest point on the A*-generated path. This ensures that the added obstacles influence the pathfinding process. We then add 10 new obstacles: 5 standard obstacles and 5 belonging to the second, more dangerous, obstacle class. As TMP contains few single-pixel obstacles, adding single-pixel obstacles from both the standard and semantic obstacle classes ensures the model distinguishes by semantics, not just by obstacle shape.

\paragraph{Waypoint following. } 
The procedure for generating the waypoint dataset is straightforward. We randomly select a cell from the set of free reachable cells that are not included in the A*-generated path, and then update its color encoding to indicate a waypoint.

\subsection{Inference Details}
All planners receive a binary occupancy grid of walkable cells to model the environment. We use octile distance as anchor heuristic $\mathbf{h_{\text{anchor}}}$. To stay consistent with the most widely used notation of minimizing the secondary heuristic, we define 

\begin{equation}
    h_C = 1 -\mathbf{C}
\end{equation}

\begin{algorithm}[t]
\caption{Comparison: Focal Search, and MHA*}
\label{alg:plannerdifferencesinone}
\begin{algorithmic}[1]
\State \textbf{Initialization:}
\State \quad Focal: OPEN $\gets$ \{start\}; CLOSED $\gets \emptyset$
\State \quad MHA*: OPEN$_{\text{anchor}}$, OPEN$_{\text{inad}} \gets$ \{start\}; CLOSED $\gets \emptyset$
\State
\While{OPEN $\neq \emptyset$} \Comment{Both Open Queues for MHA*}  
    \State \textbf{Focal:}
    \State \quad $f_{\min} \gets \min_{n' \in \text{OPEN}} \{g(n') + h_{\text{anchor}}(n')\}$
    \State \quad FOCAL $\gets \{n \in \text{OPEN} \mid g(n) + h_{\text{anchor}}(n) \leq w \cdot f_{\min}\}$
    \State \quad $n \gets \arg\min_{n' \in \text{FOCAL}} \{h_{\text{predicted}}(n')\}$
    \State
    \State \textbf{MHA*:}
    \State \quad $f_{\text{anchor}} \gets \min\{g + w_1 \cdot h_{\text{anchor}}\}$; $f_{\text{predicted}} \gets \min\{g + w_1 \cdot h_{\text{predicted}}\}$
    \State \quad \textbf{if} $g(\text{goal}) \leq \min(f_{\text{anchor}}, f_{\text{inad}})$ \textbf{then return} path
    \State \quad use\_inad $\gets (f_{\text{inad}} \leq w_2 \cdot f_{\text{anchor}})$
    \State \quad $n \gets$ top from (OPEN$_{\text{inad}}$ if use\_inad else OPEN$_{\text{anchor}}$)
    \State
    \If{$n$ is goal} \Return path 
    \EndIf
    \State Expand $n$, update OPEN and CLOSED
\EndWhile
\end{algorithmic}
\end{algorithm}

\Cref{alg:plannerdifferencesinone} provides an overview of algorithmic differences between all planners used in our experiments.

\paragraph{Focal Search. } We use Focal Search for shortest-path and waypoint experiments. The predicted soft path mask $\mathbf{C}$ serves as a secondary heuristic, where $h_{\text{predicted}} = h_C$. We use $w = 2.0$ for all shortest-path experiments relying on Focal Search (TransPath and our model).

\paragraph{Multi-Heuristic A*. } We use Multi-Heuristic A* (MHA*) for both standard obstacle avoidance and semantic obstacle avoidance. Again, we use the soft path mask $\mathbf{C}$ as a secondary heuristic, however, for MHA* we need to introduce an additional scalar $\gamma$ to compensate for the different scales of $h_{\text{anchor}}$ (distance to goal) and $h_C$ ($\in [0, 1]$). We define:
\begin{equation}
    h_{\text{predicted}} = \gamma \cdot h_C
\end{equation}
and use $\gamma = 100$ for all experiments on $64\times64$ maps. This hyperparameter requires tuning as the map resolution changes. Note that Focal Search only requires the secondary heuristic to be ordered, so that this step is redundant. For the remaining hyperparameters, we use $w_1 = 3.5$ and $w_2 = 5.0$.

\subsection{Network Architecture}

The model implements a U-Net architecture combining ResNet blocks for hierarchical feature extraction with Transformer blocks for global context modeling at the bottleneck. The network consists of three encoder stages, a four-block Transformer bottleneck, and three decoder stages with skip connections.

\paragraph{Overall configuration.}
The network operates on input tensors of shape $\mathbb{R}^{B \times 3 \times H \times W}$ and produces outputs of shape $\mathbb{R}^{{B \times 1} \times H \times W}$, where $B$ is batch size, and $H=W=64$ is the spatial resolution. All intermediate layers maintain a constant channel dimension $C = 64$.

\paragraph{Encoder path.}
Each of the three encoder stages consists of:
\begin{itemize}
    \item A ResNet block maintaining $C$ channels
    \item MaxPooling with $2 \times 2$ kernel and stride 2
    \item Resolution progression: $64 \rightarrow 32 \rightarrow 16 \rightarrow 8$
\end{itemize}

\paragraph{Transformer bottleneck.}
The bottleneck comprises four sequential Transformer blocks operating at resolution $8 \times 8$ with $C$ channels. Each block contains:
\begin{itemize}
    \item Learned positional embeddings $\mathbf{P} \in \mathbb{R}^{64 \times C}$
    \item Two cycles of multi-head self-attention with 4 heads each
    \item Three LayerNorm layers
    \item $1 \times 1$ convolution projection with residual connection
\end{itemize}

The transformation for each Transformer block is:
\begin{align}
    \mathbf{x}_0 &= \text{Reshape}(\mathbf{x}) + \mathbf{P} \in \mathbb{R}^{B \times 64 \times C} \\
    \mathbf{x}_1 &= \text{MHA}(\text{LN}(\mathbf{x}_0)) + \mathbf{x}_0 \\
    \mathbf{x}_2 &= \text{MHA}(\text{LN}(\mathbf{x}_1)) + \mathbf{x}_1 \\
    \mathbf{y} &= \text{Reshape}(\text{Conv}_{1\times1}(\text{LN}(\mathbf{x}_2)) + \mathbf{x}_2)
\end{align}

\paragraph{Decoder path.}
Each of the three decoder stages performs:
\begin{itemize}
    \item Nearest neighbor upsampling with factor 2
    \item Concatenation with encoder skip connection: $[\mathbf{x}_{\text{up}}, \mathbf{x}_{\text{skip}}] \in \mathbb{R}^{B \times 2C \times H \times W}$
    \item ResNet block reducing channels from $2C$ to $C$
    \item Resolution progression: $8 \rightarrow 16 \rightarrow 32 \rightarrow 64$
\end{itemize}

\paragraph{ResNet block.}
Each ResNet block applies two sequential convolution-normalization-activation cycles:
\begin{align}
    \mathbf{z}_1 &= \text{Dropout}(\sigma(\text{GN}(\mathbf{W}_1 * \mathbf{x})) + \mathbf{W}_{\text{skip},1} * \mathbf{x}) \\
    \mathbf{z}_2 &= \text{Dropout}(\sigma(\text{GN}(\mathbf{W}_2 * \mathbf{z}_1)) + \mathbf{W}_{\text{skip},2} * \mathbf{z}_1)
\end{align}
where $*$ denotes $3 \times 3$ convolution with padding 1, $\mathbf{W}_{\text{skip}}$ are $1 \times 1$ convolutions for channel matching, $\text{GN}$ is GroupNorm with 16 groups, $\sigma$ is SiLU activation, and Dropout has probability $p=0.1$.

\paragraph{Output layer.}
A final $1 \times 1$ convolution maps from $C$ to 1 output channel, producing the soft path mask logits.

\paragraph{Augmentations.} 
We do not apply explicit augmentations as TMP is already fully augmented.

\paragraph{Capacity and comparability.}

We chose the design and hyperparameters to closely match those of the TransPath model to ensure a fair comparison. In total, our model has slightly fewer parameters (787k) than TransPath (944k).

\subsection{Training Configuration and Hyperparameters}

\begin{table}[t!]
\centering
\small
\begin{tabular}{lcc}
\toprule
\textbf{Hyperparameter} & \textbf{Pretraining} & \textbf{Fine-tuning} \\
\midrule

\multicolumn{3}{c}{\textbf{Optimization \& Training}} \\
\midrule
Loss & WBCE ($\alpha=0.95$) & PSO-based \\
Optimizer & AdamW & AdamW \\
Learning Rate & $2.5 \cdot 10^{-5}$ & $10^{-5}$ \\
Batch size & 512 & 128 \\
Epochs & 250 & 250 \\
Opt. Steps & 250k & 250k \\
Weight decay & $\lambda=0.01$ & $\lambda=0.01$ \\
Precision & BF16 & FP32 \\
$\beta_1, \beta_2$ & 0.9, 0.999 & 0.9, 0.999 \\
$\epsilon$ & $10^{-8}$ & $10^{-8}$ \\
Grad clip & max\_norm=1.0 & max\_norm=1.0 \\
Scheduler & Cos ($\eta_{\min}=10^{-10}$) & Cos ($\eta_{\min}=10^{-8}$) \\
Seed & 1327455 & 1327455 \\

\midrule
\multicolumn{3}{c}{\textbf{Infrastructure}} \\
\midrule
GPU & 1$\times$A100 80GB & 1$\times$A100 80GB \\
CPU & AMD EPYC 7713 & AMD EPYC 7713 \\
Available Cores & 20 & 20 \\
RAM & 30GB DDR4 & 30GB DDR4 \\
Time & 10h & 7h \\
Framework & PyTorch 2.8.0 & PyTorch 2.8.0 \\

\bottomrule
\end{tabular}
\caption{Training configuration for pretraining and fine-tuning. Pretraining uses weighted BCE with $\alpha=0.95$ and a higher learning rate; fine-tuning uses only PSO-based losses with reduced learning rate and batch size at full precision.}
\label{tab:training_hyperparams}
\end{table}

\begin{table}[t!]
\centering
\small

\begin{tabular}{l l}
  \toprule
  \textbf{Reward} & \textbf{Parameters} \\
  \midrule
  $\mathcal{R}_{\mathrm{coll}}$                 & (none) \\
  $\mathcal{R}_{\mathrm{conn}}^{\mathrm{opt}}$  & $\tau^{\mathrm{opt}}_{\text{conn}} = \mathrm{lin}(8,\,16)$,\; $T = 125$ \\
  $\mathcal{R}_{\mathrm{conn}}^{\mathrm{sub}}$  & $\beta = 5.0$,\; $T = 125$ \\
  $\mathcal{R}_{\mathrm{obst}}$                 & $\tau_{\text{obs}} = 25$ \\
  $\mathcal{R}_{\mathrm{cost}}$                 & $\epsilon = 10^{-8}$ \\
  \bottomrule
\end{tabular}

\vspace{0.5em}

\begin{tabular}{l c c c c}
  \toprule
  \textbf{Technique} &
    $\lambda_{\mathrm{coll}}$ &
    $\lambda_{\mathrm{conn}}$ &
    $\lambda_{\mathrm{cost}}$ &
    $\lambda_{\mathrm{obst}}$ \\
  \midrule
  Shortest-Path       & 1.0 & 0.005 & lin(0.01,\,0.5) & --- \\
  Obstacle Avoidance  & 1.0 & 1.0 & lin(0.01,\,0.5)  & 0.2 \\
  Semantic Clearance  & 1.0 & 1.0 & lin(0.01,\,0.5)  & 0.2 \\
  Waypoint            & 1.0 & 0.005 & lin(0.01,\,0.5) & --- \\
  \bottomrule
\end{tabular}

\caption{PSO hyperparameters used in Stage~2 fine-tuning. Top: function-specific parameters. Bottom: per-objective loss weights.}
\label{tab:hyperparams}
\end{table}

\Cref{tab:training_hyperparams} lists the optimization and infrastructure setup for both training stages. Note that the fine-tuning uses sampling with replacement and one epoch refers to 1000 sampled batches.
\Cref{tab:hyperparams} reports all PSO-related hyperparameters: function-specific parameters (top) and per-objective loss weights (bottom). Objective hyperparameters that are denoted by $\mathrm{lin}(a,b)$ are linearly annealed from $a$ to $b$ over the course of fine-tuning. All hyperparameters were selected by evaluation on the validation split of TMP, final metrics were computed on the test split.

\subsection{Reduced-Budget Training Configuration}
The reduced-budget training run uses the same hyperparameters as specified in \Cref{tab:training_hyperparams,tab:hyperparams}, with the following exceptions. The pretraining is restricted to only 40 epochs with a higher learning rate of $4 \cdot 10^{-4}$. The fine-tuning uses bf16 precision, 100 instead of 250 epochs and a learning rate of $3 \cdot 10^{-4}$.

\subsection{Baseline Hyperparameters}
We retrained iA*, DAA* and Neural A* using the hyperparameters specified in the corresponding paper and Github repository, which we found to work well with our backbone architecture and dataset, too. We merely used $g_{ratio} = 0.4$ for inferencing Neural A*, as the default value of 0.5 was strongly biased towards reproducing optimal paths with only little search effort reduction (Cost Factor: 1.002, Expansion Ratio: 0.685).


\section{Extended Clearance Results}
\label{sec:ext-clear}

\subsection{Classical Cost-Map Shaping Baselines}

\paragraph{WA* with inflated obstacles. }
WA* with inflated obstacles refers to WA* which receives an occupancy grid of the environment where cells that lie within the given obstacle proximity are marked as impassable. We use a suboptimality bound of $w = 15.0$.

\paragraph{WA* with obstacle heuristic. }
WA* with obstacle heuristic refers to WA* operating on the normal occupancy grid of the environment but receives an obstacle aware heuristic of $h_0 = h_{dist} + h_W * h_{obstacle}$, where $h_{dist}$ is octile distance and $h_{obstacle}$ is the obstacle occupancy grid padded by the given clearance distance. We use $h_w = 10$ and $w = 15.0$ for the suboptimality bound. This variant is used as classical baseline in the main experiments.

\begin{table*}[t]
\centering
{\small
\setlength{\tabcolsep}{3pt}
\begin{tabular*}{\textwidth}{@{\extracolsep{\fill}} lccccc}
\toprule
& \multicolumn{2}{c}{Feasible instances} & & \multicolumn{2}{c}{Infeasible instances} \\
\cmidrule(lr){2-3}\cmidrule(lr){5-6}
Variant &
Avg.\ Dist.$\uparrow$ & Avg.\ Avoid.$\uparrow$ & &
Avg.\ Dist.$\uparrow$ & Avg.\ Avoid.$\uparrow$ \\
\midrule
\rowcolor[HTML]{E6E6E6}
\multicolumn{6}{c}{\textit{Target clearance: $d_{\min}=2$}} \\
WA* (inflated obst.)$^\dagger$     & $3.74\!\pm\!1.06$ & $(1.000\!\pm\!.000)$ & & -- & -- \\
WA* (obst.\ heuristic)             & $3.67\!\pm\!1.08$ & $0.987\!\pm\!.036$ & & $3.11\!\pm\!0.85$ & $0.946\!\pm\!.056$ \\
PT (no adaptation)                 & $3.61\!\pm\!1.21$ & $0.641\!\pm\!.133$ & & $2.87\!\pm\!1.00$ & $0.537\!\pm\!.150$ \\
PT+FT ($d_{\min}=2$)               & $\mathbf{4.15\!\pm\!0.92}$ & $\mathbf{0.998\!\pm\!.020}$ & & $\mathbf{3.55\!\pm\!0.77}$ & $\mathbf{0.954\!\pm\!.056}$ \\
\midrule
\rowcolor[HTML]{E6E6E6}
\multicolumn{6}{c}{\textit{Target clearance: $d_{\min}=3$}} \\
WA* (inflated obst.)$^\dagger$     & $4.77\!\pm\!1.14$ & $(1.000\!\pm\!.000)$ & & -- & -- \\
WA* (obst.\ heuristic)             & $4.44\!\pm\!1.13$ & $0.931\!\pm\!.112$ & & $3.47\!\pm\!0.96$ & $0.703\!\pm\!.187$ \\
PT (no adaptation)                 & $4.02\!\pm\!1.28$ & $0.574\!\pm\!.134$ & & $3.17\!\pm\!1.12$ & $0.440\!\pm\!.160$ \\
PT+FT ($d_{\min}=2$)               & $4.46\!\pm\!0.97$ & $0.679\!\pm\!.115$ & & $3.79\!\pm\!0.86$ & $0.562\!\pm\!.135$ \\
PT+FT ($d_{\min}=3$)               & $\mathbf{5.34\!\pm\!0.86}$ & $\mathbf{0.986\!\pm\!.037}$ & & $\mathbf{4.38\!\pm\!0.83}$ & $\mathbf{0.825\!\pm\!.120}$ \\
\bottomrule
\end{tabular*}
\vspace{1pt}
\raggedright
{\small \raggedright
$^\dagger$Inflating obstacles by $d_{\min}$ pixels guarantees clearance by construction (avoidance omitted), but leaves only 54.45\%/12.67\% of environments solvable at $d_{\min}{=}2/3$; reported feasible metrics cover solvable instances only, while infeasible metrics are computed on the complementary sets.\par}
}
\caption{Obstacle-clearance evaluation on TMP.
PT and PT+FT share the same Stage~1 pretrained prior.
Feasible metrics are computed on instances where target clearance $d_{\min}$ is achievable; infeasible metrics are computed on the complementary set. }
\label{tab:clearance_results_extended_tmp}
\end{table*}

\subsection{Infeasible Environment Analysis}


\Cref{tab:clearance_results_extended_tmp} separates environments according to whether the requested clearance is geometrically achievable. In addition to full clearance satisfaction, we report two graded metrics. The Average Avoidance Ratio measures the fraction of path cells that satisfy the requested clearance, while the Average Closest-Obstacle Distance averages the minimum obstacle distance attained along each path. These metrics especially capture partial improvements in environments where full satisfaction is impossible.

Inflating obstacles guarantees the requested clearance whenever a path remains available, but reduces the fraction of solvable environments to ($54.45\%$) for ($d_{\min}=2$) and ($12.67\%$) for ($d_{\min}=3$). Consequently, this baseline cannot return a path on the complementary infeasible subsets. On feasible instances, our objective-specific models achieve high average avoidance while maintaining greater obstacle distance than the obstacle-heuristic WA* baseline. On infeasible instances, where no planner can fully satisfy the requested clearance, adaptation still produces paths with substantially greater obstacle distance and higher avoidance ratios than the unadapted Stage~1 prior. For ($d_{\min}=3$), the corresponding PT+FT model also improves over obstacle-heuristic WA* from $3.47$ to $4.38$ in average closest-obstacle distance and from $0.703$ to $0.825$ in average avoidance. Thus, the learned objective provides meaningful graded behavior even when complete clearance satisfaction is geometrically impossible.

\subsection{Cross-Distribution Clearance Generalization}

\Cref{tab:clearance_results_extended} evaluates zero-shot clearance transfer to VoxelGym after training exclusively on TMP. Across both clearance targets, objective-specific fine-tuning substantially improves over the unadapted Stage~1 prior, showing that the learned clearance behavior transfers beyond the source-map distribution. For $d_{\min}=2$, PT+FT increases the average closest-obstacle distance from 3.20 to 3.90 on feasible instances while achieving an average avoidance ratio of 0.995. For $d_{\min}=3$, the corresponding objective-specific model reaches an average distance of $4.94$ and an avoidance ratio of $0.960$.

Compared with obstacle-heuristic WA*, the adapted models generally maintain a greater average distance from obstacles, although WA* achieves a higher Full Avoidance rate on feasible instances. On geometrically infeasible instances, the $d_{\min}=3$ model improves over WA* in both average closest-obstacle distance (4.46 versus 3.65) and average avoidance (0.871 versus 0.851). These results indicate that the adaptation does not merely memorize TMP-specific obstacle configurations: it transfers a graded clearance preference to an unseen map distribution. Nevertheless, the lower Full Avoidance rates relative to WA* on feasible VoxelGym instances show that this transfer is not uniformly superior across all metrics.

\begin{table*}[t]
\centering
{\small
\setlength{\tabcolsep}{3pt}
\begin{tabular*}{\textwidth}{@{\extracolsep{\fill}} lcccccc}
\toprule
& \multicolumn{3}{c}{Feasible instances} & & \multicolumn{2}{c}{Infeasible instances} \\
\cmidrule(lr){2-4}\cmidrule(lr){6-7}
Variant &
Avg.\ Dist.$\uparrow$ & Avg.\ Avoid.$\uparrow$ & Full Avoid.$\uparrow$ & &
Avg.\ Dist.$\uparrow$ & Avg.\ Avoid.$\uparrow$ \\
\midrule
\rowcolor[HTML]{E6E6E6}
\multicolumn{7}{c}{\textit{Target clearance: $d_{\min}=2$}} \\
WA* (inflated obst.)$^\dagger$     & $3.44\!\pm\!0.90$ & $(1.000\!\pm\!.000)$ & $(1.000)$ & & -- & -- \\
WA* (obst.\ heuristic)             & $3.43\!\pm\!0.91$ & $0.993\!\pm\!.045$ & $\mathbf{0.970}$ & & $3.14\!\pm\!0.80$ & $\mathbf{0.947\!\pm\!.092}$ \\
PT (no adaptation)                 & $3.20\!\pm\!1.14$ & $0.589\!\pm\!.165$ & $0.000$ & & $2.99\!\pm\!1.08$ & $0.555\!\pm\!.162$ \\
PT+FT ($d_{\min}=2$)               & $\mathbf{3.90\!\pm\!1.01}$ & $\mathbf{0.995\!\pm\!.028}$ & $0.924$ & & $\mathbf{3.72\!\pm\!1.01}$ & $0.946\!\pm\!.086$ \\
\midrule
\rowcolor[HTML]{E6E6E6}
\multicolumn{7}{c}{\textit{Target clearance: $d_{\min}=3$}} \\
WA* (inflated obst.)$^\dagger$     & $4.28\!\pm\!0.81$ & $(1.000\!\pm\!.000)$ & $(1.000)$ & & -- & -- \\
WA* (obst.\ heuristic)             & $4.15\!\pm\!0.84$ & $0.953\!\pm\!.136$ & $\mathbf{0.859}$ & & $3.65\!\pm\!0.81$ & $0.851\!\pm\!.187$ \\
PT (no adaptation)                 & $3.42\!\pm\!1.11$ & $0.503\!\pm\!.163$ & $0.000$ & & $3.00\!\pm\!1.11$ & $0.403\!\pm\!.186$ \\
PT+FT ($d_{\min}=2$)               & $4.08\!\pm\!1.00$ & $0.628\!\pm\!.176$ & $0.004$ & & $3.720\!\pm\!0.987$ & $0.543\!\pm\!.185$ \\
PT+FT ($d_{\min}=3$)               & $\mathbf{4.94\!\pm\!0.91}$ & $\mathbf{0.960\!\pm\!.091}$ & $0.630$ & & $\mathbf{4.46\!\pm\!0.94}$ & $\mathbf{0.871\!\pm\!.144}$ \\
\bottomrule
\end{tabular*}
\vspace{1pt}
\raggedright
{\small \raggedright
$^\dagger$Inflating obstacles by $d_{\min}$ pixels guarantees clearance by construction (avoidance omitted), but leaves only 85.80\%/41.96\% of environments solvable at $d_{\min}{=}2/3$; reported feasible metrics cover solvable instances only, while infeasible metrics are computed on the complementary sets.\par}
}
\caption{Cross-distribution obstacle-clearance evaluation on VoxelGym~50k. All learned models are trained on TMP~640k and evaluated without retraining. Feasible/infeasible splits are based on whether the target clearance $d_{\min}$ is geometrically achievable. }
\label{tab:clearance_results_extended}
\end{table*}


\section{Extended Adaptation Results}
\label{sec:ext-adapt}

\subsection{Classical Cost-Map Shaping Baselines}

\begin{table*}[t]
\centering
{\small
\setlength{\tabcolsep}{3pt}
\begin{tabular*}{\textwidth}{@{\extracolsep{\fill}} lccccc}
\toprule
Variant & Exp.\ $\downarrow$ & Full Avoid.$\uparrow$ & Class-conditioned Full Avoid.$\uparrow$ & Avg.\ Avoid.$\uparrow$ & Avg.\ Dist.$\uparrow$ \\
\midrule
WA* (obst.\ heuristic)    & $0.511\!\pm\!.470$ & $0.529$ & $0.660$ & $0.936\!\pm\!.073$ & $3.22\!\pm\!0.79$ \\
PT+FT ($d_{\min}=2$)        & $0.216\!\pm\!.392$ & $0.812$ & $0.143$ & $0.970\!\pm\!.052$ & $3.88\!\pm\!0.75$ \\
PT+FT (class cond. avoidance)        & $\mathbf{0.188\!\pm\!.286}$ & $\mathbf{0.852}$ & $\mathbf{0.684}$ & $\mathbf{0.971\!\pm\!.050}$ & $\mathbf{3.99\!\pm\!0.85}$ \\
\bottomrule
\end{tabular*}
}
\caption{Class-conditioned obstacle clearance evaluation on TMP. PT+FT (class cond. avoidance) is fine-tuned with the class conditioned clearance loss (\Cref{eq:loss_clearance}); PT+FT ($d_{\min}=2$) uses the standard clearance loss without semantic distinction. Full Avoidance and Class-conditioned Avoidance are computed on feasible instances.}
\label{tab:semantic_avoidance_results}
\end{table*}

\Cref{tab:semantic_avoidance_results} compares class-conditioned clearance against a classical planner and a standard obstacle clearance fine-tuned model. The classical WA* baseline treats all obstacles uniformly, achieving moderate class-conditioned avoidance but at over twice the search effort. PT+FT ($d_{\min}=2$), fine-tuned without obstacle class distinction, achieves strong overall avoidance but only 14.3\% full class-conditioned clearance, since it cannot differentiate obstacle classes. PT+FT (class cond.\ avoid.), trained with the class--aware loss, reaches the highest class-conditioned avoidance (68.4\%) while maintaining comparable overall clearance, confirming that the PSO framework can encode class-specific distance preferences.

\subsection{Fine-tuning Robustness and Two-Stage Necessity}

\begin{table}[t]
\centering
{\small
\setlength{\tabcolsep}{3pt}
\begin{tabular*}{\columnwidth}{@{\extracolsep{\fill}} lcccc}
\toprule
Method
& Cost $\downarrow$
& Exp.\ $\downarrow$
& H.\ Val.\ $\uparrow$
& Opt.\ F.\ $\uparrow$ \\
\midrule
FT-only & $1.004\!\pm\!.01$ & $0.168\!\pm\!.11$ & $0.986$ & $0.795$ \\
\midrule
PT+FT & $1.002\!\pm\!.01$ & $0.163\!\pm\!.11$ & $0.991$ & $0.886$ \\
Bidirectional & $1.002\!\pm\!.01$ & $0.162\!\pm\!.10$ & $0.992$ & $0.884$ \\
\midrule
PT+FT (Bin. occ.) & $1.002{\pm}.01$ & $0.163{\pm}.11$ & $0.992$ & $0.883$ \\
\bottomrule
\end{tabular*}}
\caption{Prior stability on shortest-path planning and comparison of input representation. FT-only achieves feasible paths but with notably lower optimality. PT+FT and bidirectional adaptation (obstacle avoidance → shortest-paths) perform comparably. PT+FT (Bin. occ.) uses binary occupancy inputs instead of RGB. All models trained on TMP 640k.}
\label{tab:bidirectional_optimality}
\end{table}

\begin{table*}[t]
\centering
{\small
\setlength{\tabcolsep}{3pt}
\begin{tabular*}{\textwidth}{@{\extracolsep{\fill}} lcccccc}
\toprule
& & \multicolumn{3}{c}{Feasible instances} & \multicolumn{2}{c}{Infeasible instances} \\
\cmidrule(lr){3-5}\cmidrule(lr){6-7}
Variant & Exp.\ $\downarrow$ &
Avg.\ Dist.$\uparrow$ & Avg.\ Avoid.$\uparrow$ & Full Avoid.$\uparrow$ &
Avg.\ Dist.$\uparrow$ & Avg.\ Avoid.$\uparrow$ \\
\midrule
\rowcolor[HTML]{E6E6E6}
\multicolumn{7}{c}{\textit{Target clearance: $d_{\min}=2$}} \\
PT+FT               & $0.163\!\pm\!.226$ & $4.15\!\pm\!0.92$ & $0.998\!\pm\!.020$ & $0.962$ & $3.55\!\pm\!0.77$ & $0.954\!\pm\!.056$ \\
Bidirectional     & $0.172\!\pm\!.232$ & $4.51\!\pm\!1.00$ & $0.998\!\pm\!.019$ & $0.969$ & $3.83\!\pm\!0.84$ & $0.955\!\pm\!.055$ \\
\bottomrule
\end{tabular*}}
\caption{Prior stability on obstacle clearance. PT+FT and bidirectional adaptation (shortest-paths → obstacle avoidance) perform comparably. FT-only collapsed during training and is omitted. Evaluated on TMP with $d_{\min}=2$; feasible/infeasible splits reflect geometric achievability.}
\label{tab:clearance_results_bidirectional}
\end{table*}

We test whether an already-fine-tuned prior can be re-fine-tuned for a different objective (bidirectional adaptation). \Cref{tab:bidirectional_optimality,tab:clearance_results_bidirectional} show that bidirectional models match their PT+FT counterparts on both shortest-path and obstacle clearance metrics, confirming that fine-tuning does not irreversibly overwrite the prior. We additionally include an FT-only baseline (no pretraining) in \Cref{tab:bidirectional_optimality}; on obstacle clearance, FT-only training collapsed entirely, producing no viable paths, underscoring the necessity of the two-stage pipeline.

\subsection{Bidirectional Adaptation Experiments}

We analyze the robustness of the fine-tuning by adapting an already fine-tuned prior again for a different objective (bidirectional). \Cref{tab:bidirectional_optimality} shows results for the model that was fine-tuned on obstacle avoidance first and then fine-tuned for shortest-paths and compares results to the model that was only fine-tuned for this objective. \Cref{tab:clearance_results_bidirectional} shows the same for obstacle avoidance, too. Both experiments indicate that the fine-tuning stage is robust enough to revert and modify previously learned preference.


\section{Additional Ablation Studies}
\label{sec:ablations}

\subsection{Input Representation}

We ablate our RGB encoded input representation against normal binary occupancy representations. After the same Stage 1 pretraining and Stage 2 fine-tuning, both models produce near-identical results (see \Cref{tab:bidirectional_optimality}), verifying that the input representation is not responsible for the gains compared to the baselines. Therefore, the role of our RGB encoding is solely to provide the flexibility of adding new classes after the pretraining without retraining the full model from scratch.

\subsection{Alpha Value for Hard Validity}

\begin{figure}
    \centering
    \includegraphics[width=1\linewidth]{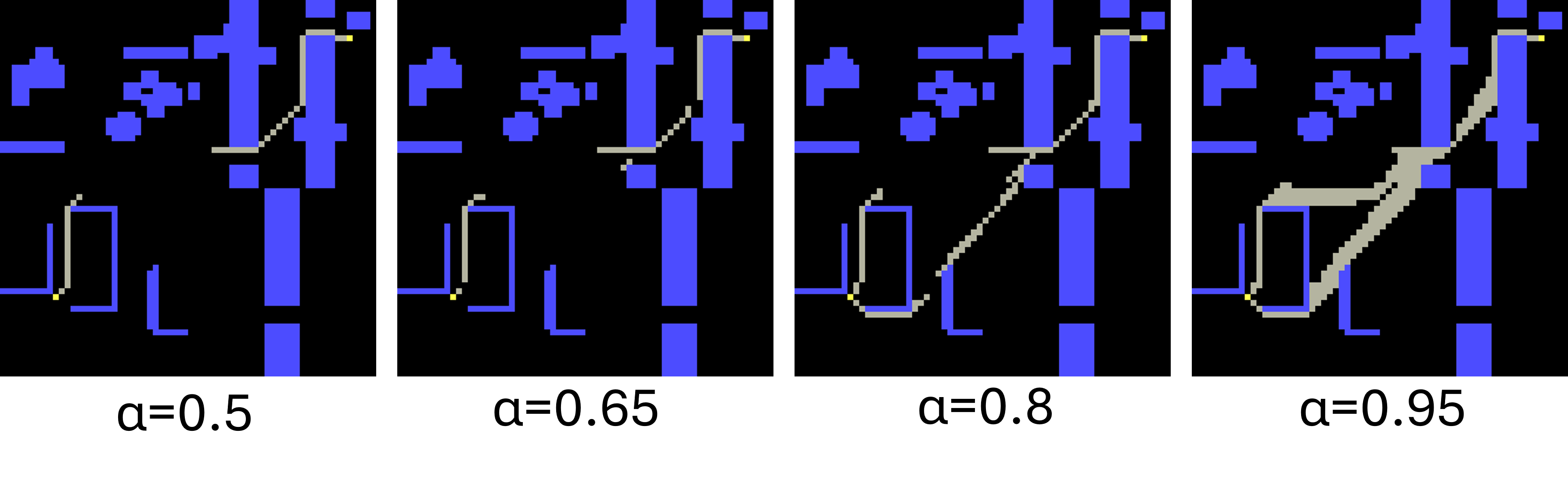}
    \caption{Effect of $\alpha$ on the Stage 1 prior. Higher $\alpha$ increases recall and preserves connectivity for Stage 2. Lower values cause fragmentation. }
    \label{fig:alpha_imgs}
\end{figure}

\Cref{tab:alpha_tmp_results} shows quantitative results for models pretrained with different $\alpha$ values, and \Cref{fig:alpha_imgs} shows qualitative results. Higher $\alpha$ increases recall, producing wider predicted regions that preserve path connectivity. This becomes a stable foundation for Stage~2 adaptation. Lower values lead to fragmented predictions where the fine-tuning stage has no connected structure to refine.

\begin{table}[t]
\centering
{\small
\setlength{\tabcolsep}{3pt}
\begin{tabular*}{\columnwidth}{@{\extracolsep{\fill}} lcccc}
\toprule
\small{Method}
& Cost $\downarrow$
& Exp.\ $\downarrow$
& H.\ Val.\ $\uparrow$
& Opt.\ F.\ $\uparrow$ \\
\midrule
PT($\alpha=0.5$) & $1.010\!\pm\!.01$ & $0.160\!\pm\!.13$ & $0.080$ & $0.499$ \\
PT($\alpha=0.65$) & $1.010\!\pm\!.01$ & $0.160\!\pm\!.13$ & $0.214$ & $\mathbf{0.522}$ \\
PT($\alpha=0.8$) & $1.010\!\pm\!.01$ & $0.161\!\pm\!.13$ & $0.576$ & $0.498$ \\
\midrule \\
PT($\alpha=0.95$) & $1.010\!\pm\!.01$ & $0.167\!\pm\!.15$ & $\mathbf{0.984}$ & $0.470$ \\

\bottomrule
\end{tabular*}}
\caption{Quantitative results for PT models trained with different $\alpha$ values.
All models were trained on TMP~640k. }
\label{tab:alpha_tmp_results}
\end{table}


\subsection{BCE Prior Instability on Non-Optimal Objectives}

\begin{figure}
    \centering
    \includegraphics[width=1\linewidth]{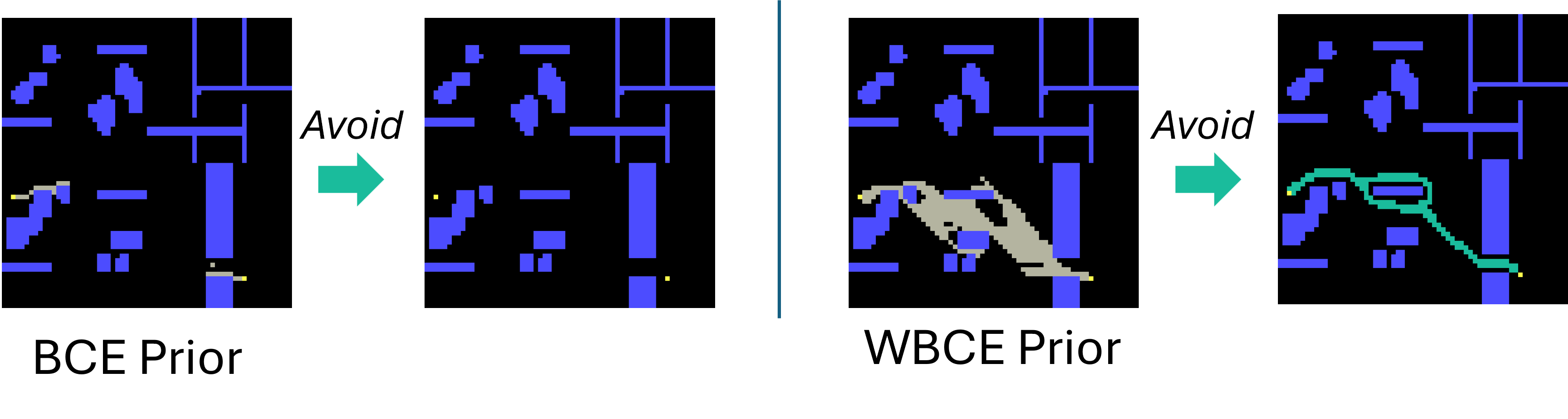}
    \caption{BCE (left) vs.\ WBCE (right) prior stability on one-tenth of TMP. The BCE-pretrained model collapses during clearance fine-tuning. WBCE remains stable.}
    \label{fig:instability}
\end{figure}

We assess the impact of the loss function by comparing a BCE-pretrained prior with a WBCE-pretrained prior on obstacle avoidance. Trained on the large TMP 640k dataset, both models yield comparable metrics. In contrast, using a smaller dataset (one-tenth of TMP for this experiment) reveals instability and collapse for the BCE-pretrained model (see \Cref{fig:instability}), whereas the WBCE-pretrained model maintains robust obstacle avoidance. Across metrics, the two pretrained models diverge mainly in Hard Validity, underscoring its importance when training under non-optimal objectives.

\subsection{Effect of PSO Components and Weights}


We analyze the importance of individual PSOs by replacing one component at a time. In addition, we show stability across hyperparameter choices and give further insight into the role that each PSO plays.

\paragraph{Connectivity.}

\begin{figure}
    \centering
    \includegraphics[width=1\linewidth]{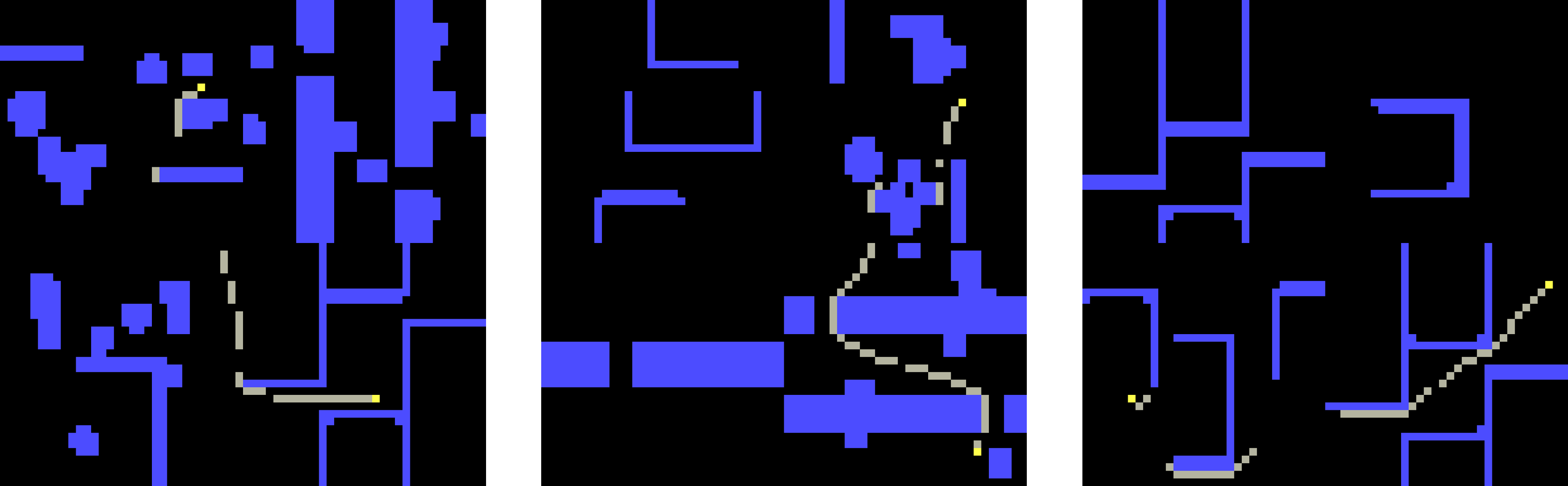}
    \caption{Paths predictions generated by the model trained without the connectivity PSO. Without this objective, predictions fragment and search frequently fails. }
    \label{fig:noconn}
\end{figure}

\Cref{fig:noconn} shows that removing the connectivity PSO leads to highly disconnected paths. 

\paragraph{Collision.} 

Removing the collision PSO leads to an increase of collision rate from $0\%$ to $0.47\%$ on TMP. Since no other PSO components encourage collisions the increase is rather small, yet this component is important to minimize collision risk.

\paragraph{Cost.}

\begin{figure}
    \centering
    \includegraphics[width=1\linewidth]{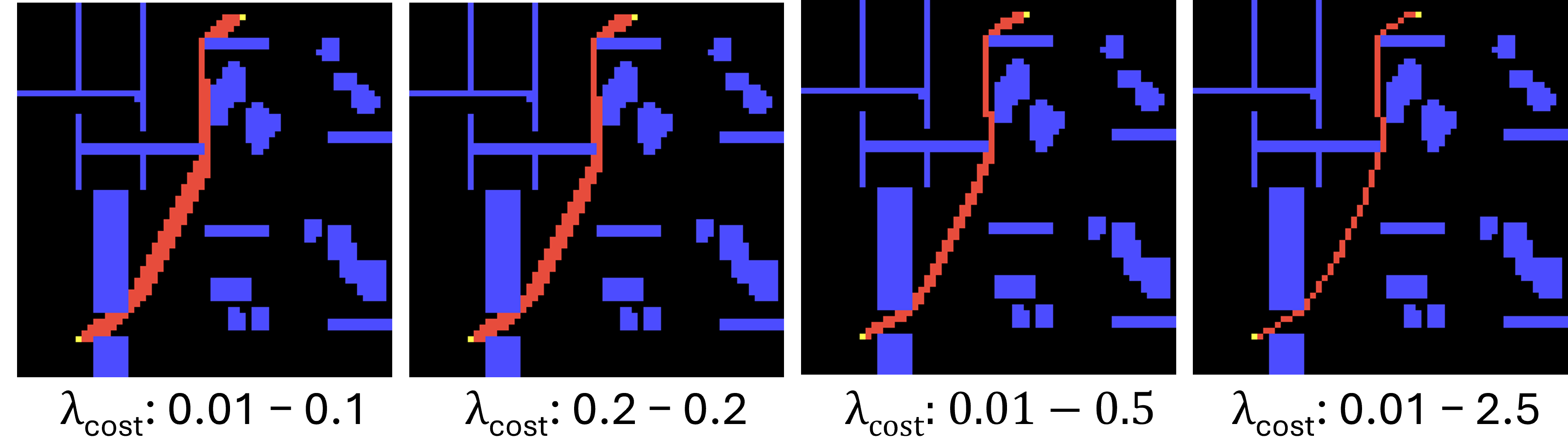}
    \caption{Predictions of models fine-tuned with different $\lambda_{cost}$. Higher values produce narrower paths, reducing search effort at the expense of optimality.}
    \label{fig:costpso_imgs}
\end{figure}

\Cref{fig:costpso_imgs} shows that the cost PSO regulates path thickness, and \Cref{tab:lambdacost} confirms that increasing $\lambda_{\text{cost}}$ reduces search effort at the expense of slightly less optimal paths, showing that $\lambda_{\text{cost}}$ acts as a natural tradeoff between optimality and efficiency. We use $\lambda_{\text{cost}}: 0.01-0.5$ scheduling throughout all shortest-path experiments for a balanced optimality/efficiency tradeoff.

\begin{table}[t]
\centering
{\small
\setlength{\tabcolsep}{3pt}
\begin{tabular*}{\columnwidth}{@{\extracolsep{\fill}} lcccc}
\toprule
\small{Method}
& Cost $\downarrow$
& Exp.\ $\downarrow$
& H.\ Val.\ $\uparrow$
& Opt.\ F.\ $\uparrow$ \\
\midrule
$\lambda_{cost}$: 0.01-0.1 & $1.001\!\pm\!.01$ & $0.233\!\pm\!.13$ & $0.995$ & $0.925$ \\
$\lambda_{cost}$: 0.2-0.2 & $1.002\!\pm\!.01$ & $0.202\!\pm\!.12$ & $0.994$ & $0.908$ \\
$\lambda_{cost}$: 0.01-0.5 & $1.002\!\pm\!.01$ & $0.163\!\pm\!.11$ & $0.991$ & $0.886$ \\
$\lambda_{cost}$: 0.01-2.5 & $1.003\!\pm\!.01$ & $0.121\!\pm\!.09$ & $0.976$ & $0.824$ \\


\bottomrule
\end{tabular*}}
\caption{Quantitative results for PT+FT models trained with different $\lambda_{cost}$ schedules.
All models were trained on TMP~640k. }
\label{tab:lambdacost}
\end{table}


\section{Extended Shortest-Path Results}
\label{sec:ext-sp}

\subsection{VoxelGym as Training Source}

\Cref{tab:voxelgym_results} reports shortest-path results when VoxelGym is used as the training source instead of TMP. Because retraining every baseline on a new dataset is computationally expensive, we restrict this experiment to Neural A*, which serves as a representative differentiable planning baseline. Stage~1 pretraining alone produces a connected prior but yields limited optimal-path recovery and high search effort. Applying Stage~2 fine-tuning substantially improves all metrics: the full pipeline achieves a Cost Factor of 1.002, an Expansion Ratio of 0.255, a Hard Validity of 0.990, and an Optimal Found Ratio of 0.937. It thereby outperforms Neural A* in both path quality and search efficiency. These results show that the benefit of the two-stage training procedure is not specific to TMP and remains effective when VoxelGym is used as the source distribution. Because the comparison includes only one baseline, we interpret this experiment as a robustness check of the training procedure rather than a comprehensive benchmark on VoxelGym.

\begin{table}[t]
\centering
{\small
\setlength{\tabcolsep}{3pt}
\begin{tabular*}{\columnwidth}{@{\extracolsep{\fill}} lcccc}
\toprule
\small{Method}
& Cost $\downarrow$
& Exp.\ $\downarrow$
& H.\ Val.\ $\uparrow$
& Opt.\ F.\ $\uparrow$ \\
\midrule
Neural A*
  & $1.005\!\pm\!.01$ & $0.682\!\pm\!.19$ & -- & $0.764$ \\
Ours (PT)
  & $1.019\!\pm\!.02$ & $1.419\!\pm\!.89$ & $0.907$ & $0.377$ \\
Ours (PT+FT)
  & $\mathbf{1.002\!\pm\!.01}$ & $\mathbf{0.255\!\pm\!.13}$ & $\mathbf{0.990}$ & $\mathbf{0.937}$ \\
\bottomrule
\end{tabular*}}
\caption{Shortest-path results with VoxelGym as the training source (cf.\ TMP as primary source in the main paper). PT: Stage~1 only, PT+FT: full pipeline.}
\label{tab:voxelgym_results}
\end{table}

\subsection{Per-Dataset Qualitative Results}

\begin{figure*}[t]
    \centering
    \includegraphics[width=1\linewidth]{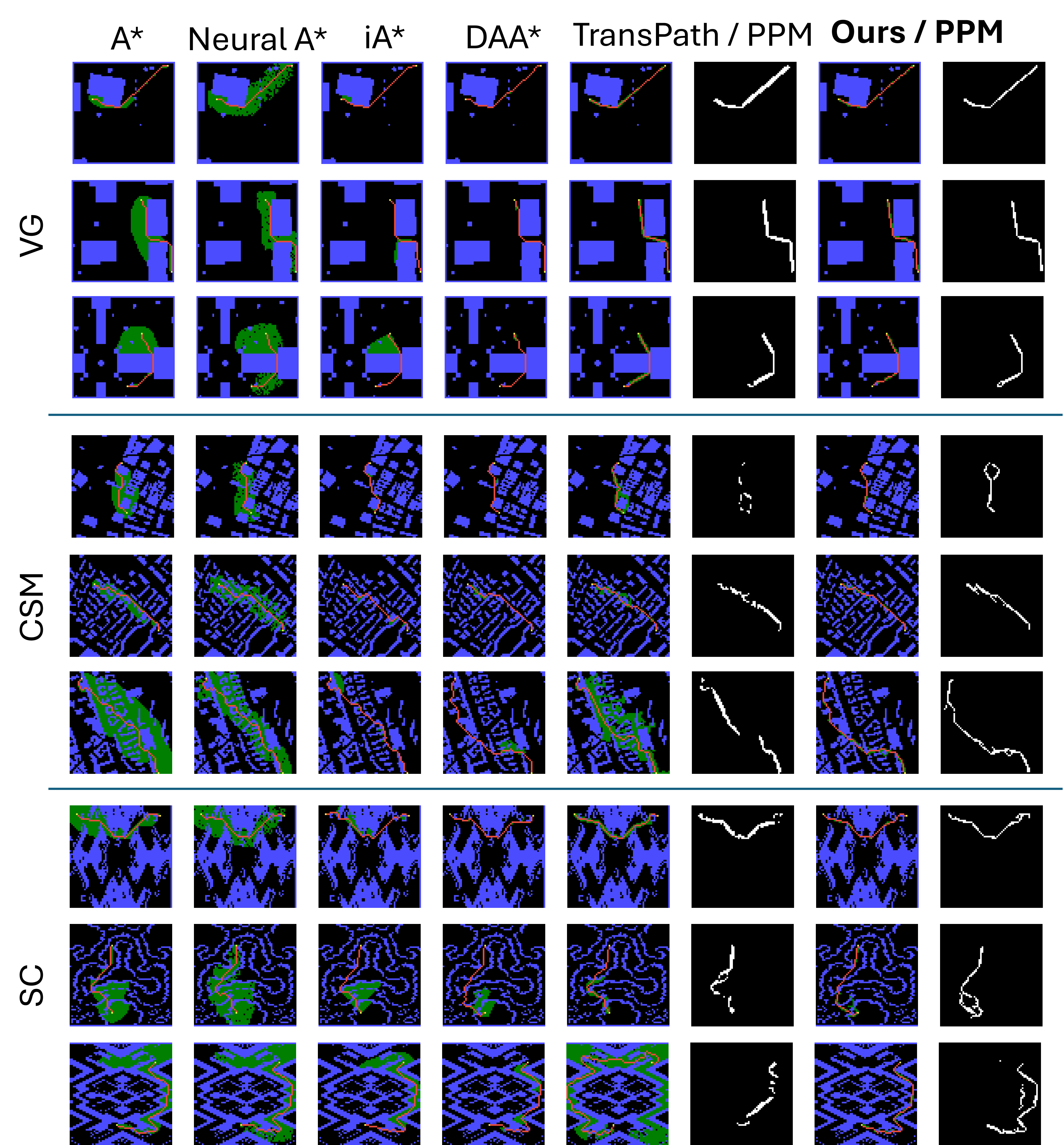}
    \caption{Selected examples for the shortest-path objective on evaluation-only datasets. Expanded nodes are shown in green, and the resulting path is shown in red.
        PPMs shown only for methods that produce them.}
    \label{fig:ood_qualitative}
\end{figure*}

\Cref{fig:ood_qualitative} shows demonstrations of our planner and comparisons with baselines on the out-of-distribution datasets.


\section{Additional Qualitative Visualizations}
\label{sec:qual}
\begin{figure*}
    \centering
    \includegraphics[width=\linewidth]{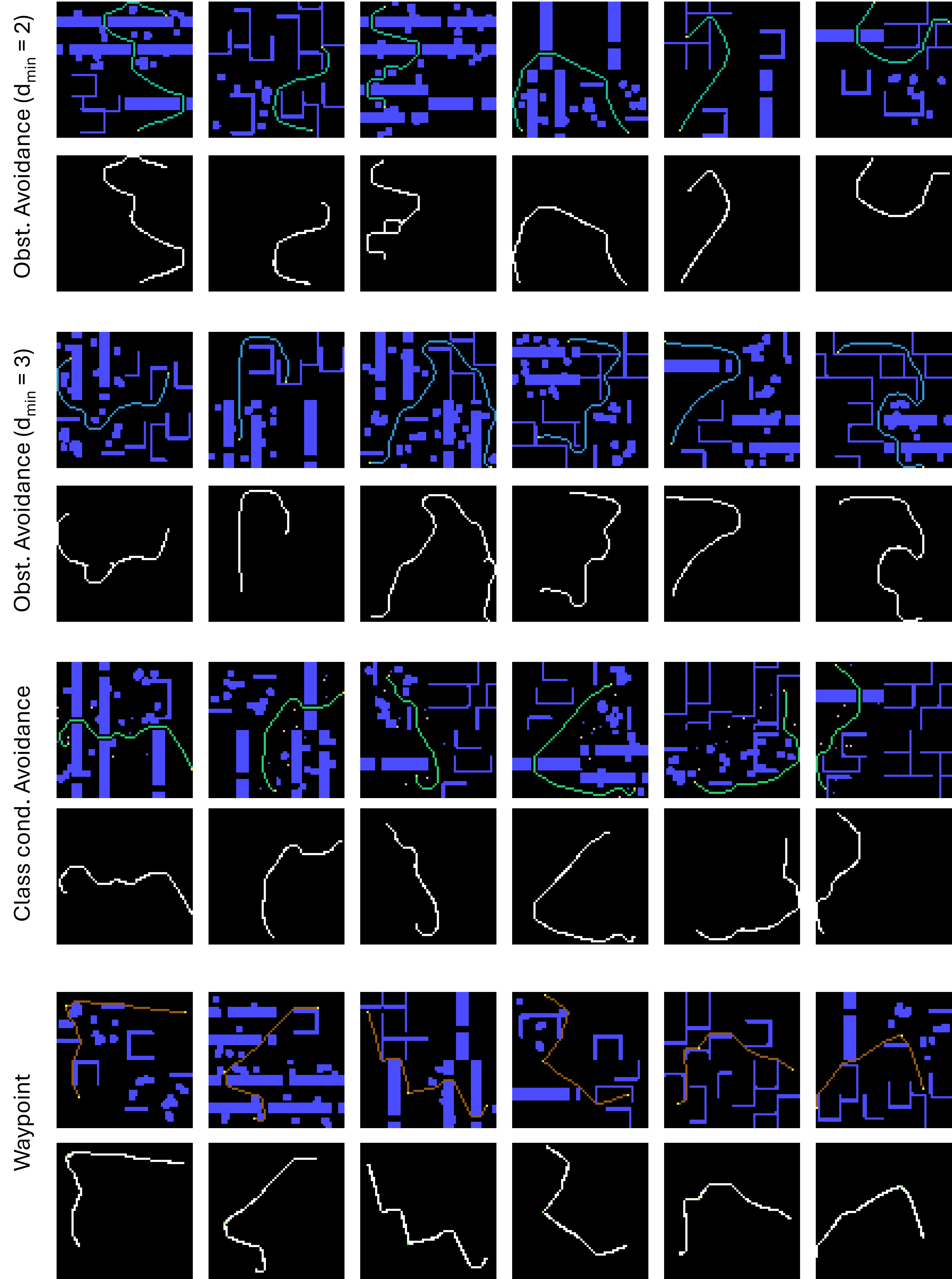}
    \caption{Additional qualitative results across all non-optimal objectives. From top to bottom: obstacle avoidance ($d_{\min}=2$), obstacle avoidance ($d_{\min}=3$), class-conditioned clearance, and waypoint following.}
    \label{fig:moreimages}
\end{figure*}

\Cref{fig:moreimages} presents additional qualitative results for all non-optimal objectives. Across obstacle avoidance at both clearance levels, the model consistently routes paths away from obstacles while maintaining connectivity. Class-conditioned clearance correctly distinguishes classes, granting more clearance to semantic obstacles. Waypoint following shows that the model produces paths that pass through the designated waypoint without unnecessary detours.

\end{document}